\pdfoutput=1
\documentclass{article} 
\usepackage[T1]{fontenc}
\usepackage{iclr2027_conference,times}

\usepackage{amsmath,amsfonts,bm}

\def\eqref#1{equation~\ref{#1}}

\def\1{\bm{1}}

\DeclareMathAlphabet{\mathsfit}{\encodingdefault}{\sfdefault}{m}{sl}
\SetMathAlphabet{\mathsfit}{bold}{\encodingdefault}{\sfdefault}{bx}{n}

\newcommand{\R}{\mathbb{R}}

\DeclareMathOperator*{\argmax}{arg\,max}

\usepackage{xcolor}
\definecolor{linkblue}{RGB}{20,70,160}
\usepackage{colortbl}
\definecolor{rkbest}{RGB}{200,30,30}
\definecolor{rksecond}{RGB}{28,82,168}
\definecolor{rkthird}{RGB}{110,110,110}
\definecolor{ourfill}{RGB}{233,240,250}
\newcommand{\best}[1]{\textcolor{rkbest}{\textbf{#1}}}
\newcommand{\second}[1]{\textcolor{rksecond}{#1}}
\newcommand{\third}[1]{\textcolor{rkthird}{#1}}
\usepackage[colorlinks=true,linkcolor=linkblue,citecolor=linkblue,urlcolor=linkblue,
            filecolor=linkblue]{hyperref}
\usepackage{url}
\usepackage{amsmath,amssymb}
\usepackage{amsthm}

\usepackage{booktabs}
\usepackage{tabularx}
\newcolumntype{Y}{>{\centering\arraybackslash}X}
\usepackage{longtable}
\makeatletter
\def\LT@makecaption#1#2#3{%
  \LT@mcol\LT@cols c{\hbox to\z@{\hss\parbox[t]\LTcapwidth{%
    \reset@font\normalsize
    \sbox\@tempboxa{#1{#2: }#3}%
    \ifdim\wd\@tempboxa>\hsize
      #1{#2: }#3%
    \else
      \hbox to\hsize{\hfil\box\@tempboxa\hfil}%
    \fi
    \endgraf\vskip\baselineskip}%
  \hss}}}
\makeatother
\usepackage{pifont}
\usepackage{enumitem}
\setlist{itemsep=1pt,parsep=0pt,topsep=3pt,partopsep=0pt}
\newcommand{\yes}{\ding{51}}
\newcommand{\no}{\textcolor{black!35}{\ding{55}}}
\newcommand{\apptab}{\centering\scriptsize\setlength{\tabcolsep}{2.6pt}%
  \renewcommand{\arraystretch}{1.0}}
\newcommand{\PAR}[1]{\noindent\textbf{#1}~}
\usepackage{graphicx}
\usepackage{multirow}
\usepackage{capt-of}

\newcommand{\Sph}{\mathbb{S}}
\newcommand{\Csph}{C}

\newcommand{\STscore}{\mathcal{R}}

\newcommand{\fhat}{\hat{f}}
\newcommand{\Fcon}{F}
\newcommand{\Ebd}{E}



\def\lyrNimg{1{,}970}
\def\lyrNcand{20{,}270}
\def\lyrDiceThree{.159}

\def\lyrDiceOneTwo{.554}

\def\bbBdice{.574}
\def\bbBauroc{.937}

\def\bbLauroc{.959}
\def\bbSdice{.256}
\def\bbSauroc{.747}

\def\bbNcand{11{,}848}

\def\nsRnd{.758}
\def\nsConf{.769}
\def\nsHeu{.773}
\def\nsSt{.784}
\def\nsOrc{.818}

\def\nsNds{4}
\def\nsNimg{913}

\def\fpSam{.276}

\def\fpOrc{.685}

\def\nDatasets{15}

\def\nsFinalDelta{+.011}
\def\nsFinalCi{[+.006,\,+.017]}
\def\nsNimg{913}

\def\ablMeta{1{,}157 images, 11{,}848 candidates}
\def\ablPoolsRange{+.099--+.158}
\def\ablPoolsN{all four}

\def\ablE{.469}

\def\ablFull{.629}

\def\ablSamE{.433}

\def\ablTokE{.579}

\def\ablPcE{.595}

\def\ablDropF{$-$.126}
\def\ablDropC{$-$.038}
\def\ablDropE{$-$.096}

\def\ablDHeu{+.038}
\def\ablDE{+.161}

\def\ablCiHeu{[+.023,\,+.052]}
\def\ablCiE{[+.140,\,+.182]}

\def\xaPairCCod{0.851}
\def\xaPairACod{0.481}

\def\xaPairCDuts{0.815}
\def\xaPairADuts{0.491}

\def\xaPairCDis{0.791}
\def\xaPairADis{0.537}

\def\xmFrameMin{9\%}
\def\xmFrameMax{16\%}

\def\pfSelRandomCod{.1319}
\def\pfSelOracleCod{.6770}

\def\pfSelNCod{1979}
\def\pfSelKCod{10.2}
\def\pfTrnRandomCod{.6780}
\def\pfTrnOracleCod{.8517}

\def\pfTrnNCod{3857}
\def\pfTrnKCod{25.6}
\def\pfSelRandomSod{.1697}
\def\pfSelOracleSod{.7878}

\def\pfSelNSod{4993}
\def\pfSelKSod{10.7}
\def\pfTrnRandomSod{.7709}
\def\pfTrnOracleSod{.9486}

\def\pfTrnNSod{4025}
\def\pfTrnKSod{25.0}
\def\pfSelRandomDis{.0992}
\def\pfSelOracleDis{.4856}

\def\pfSelNDis{465}
\def\pfSelKDis{11.3}
\def\pfTrnRandomDis{.5762}
\def\pfTrnOracleDis{.8385}

\def\pfTrnNDis{2969}
\def\pfTrnKDis{28.4}
\def\pfSelRandomLl{.1307}
\def\pfSelOracleLl{.6028}

\def\pfSelNLl{1932}
\def\pfSelKLl{9.4}

\def\pfSelRandMin{.0992}
\def\pfSelRandMax{.1697}

\providecommand{\ablCiE}{\textbf{[ablCiE not synced]}}
\providecommand{\ablCiHeu}{\textbf{[ablCiHeu not synced]}}

\providecommand{\ablDE}{\textbf{[ablDE not synced]}}
\providecommand{\ablDHeu}{\textbf{[ablDHeu not synced]}}

\providecommand{\ablDropC}{\textbf{[ablDropC not synced]}}
\providecommand{\ablDropE}{\textbf{[ablDropE not synced]}}
\providecommand{\ablDropF}{\textbf{[ablDropF not synced]}}
\providecommand{\ablE}{\textbf{[ablE not synced]}}

\providecommand{\ablFull}{\textbf{[ablFull not synced]}}

\providecommand{\ablMeta}{\textbf{[ablMeta not synced]}}

\providecommand{\ablPcE}{\textbf{[ablPcE not synced]}}

\providecommand{\ablPoolsN}{\textbf{[ablPoolsN not synced]}}
\providecommand{\ablPoolsRange}{\textbf{[ablPoolsRange not synced]}}

\providecommand{\ablSamE}{\textbf{[ablSamE not synced]}}

\providecommand{\ablTokE}{\textbf{[ablTokE not synced]}}

\providecommand{\bbBauroc}{\textbf{[bbBauroc not synced]}}
\providecommand{\bbBdice}{\textbf{[bbBdice not synced]}}
\providecommand{\bbLauroc}{\textbf{[bbLauroc not synced]}}

\providecommand{\bbNcand}{\textbf{[bbNcand not synced]}}

\providecommand{\bbSauroc}{\textbf{[bbSauroc not synced]}}
\providecommand{\bbSdice}{\textbf{[bbSdice not synced]}}

\providecommand{\fpOrc}{\textbf{[fpOrc not synced]}}

\providecommand{\fpSam}{\textbf{[fpSam not synced]}}

\providecommand{\lyrDiceOneTwo}{\textbf{[lyrDiceOneTwo not synced]}}

\providecommand{\lyrDiceThree}{\textbf{[lyrDiceThree not synced]}}

\providecommand{\lyrNcand}{\textbf{[lyrNcand not synced]}}
\providecommand{\lyrNimg}{\textbf{[lyrNimg not synced]}}

\providecommand{\nDatasets}{\textbf{[nDatasets not synced]}}

\providecommand{\nsConf}{\textbf{[nsConf not synced]}}

\providecommand{\nsFinalCi}{\textbf{[nsFinalCi not synced]}}
\providecommand{\nsFinalDelta}{\textbf{[nsFinalDelta not synced]}}

\providecommand{\nsHeu}{\textbf{[nsHeu not synced]}}

\providecommand{\nsNds}{\textbf{[nsNds not synced]}}
\providecommand{\nsNimg}{\textbf{[nsNimg not synced]}}
\providecommand{\nsOrc}{\textbf{[nsOrc not synced]}}

\providecommand{\nsRnd}{\textbf{[nsRnd not synced]}}

\providecommand{\nsSt}{\textbf{[nsSt not synced]}}

\providecommand{\pfSelKCod}{\textbf{[pfSelKCod not synced]}}
\providecommand{\pfSelKDis}{\textbf{[pfSelKDis not synced]}}
\providecommand{\pfSelKLl}{\textbf{[pfSelKLl not synced]}}
\providecommand{\pfSelKSod}{\textbf{[pfSelKSod not synced]}}

\providecommand{\pfSelNCod}{\textbf{[pfSelNCod not synced]}}
\providecommand{\pfSelNDis}{\textbf{[pfSelNDis not synced]}}
\providecommand{\pfSelNLl}{\textbf{[pfSelNLl not synced]}}
\providecommand{\pfSelNSod}{\textbf{[pfSelNSod not synced]}}

\providecommand{\pfSelOracleCod}{\textbf{[pfSelOracleCod not synced]}}
\providecommand{\pfSelOracleDis}{\textbf{[pfSelOracleDis not synced]}}
\providecommand{\pfSelOracleLl}{\textbf{[pfSelOracleLl not synced]}}
\providecommand{\pfSelOracleSod}{\textbf{[pfSelOracleSod not synced]}}

\providecommand{\pfSelRandMax}{\textbf{[pfSelRandMax not synced]}}
\providecommand{\pfSelRandMin}{\textbf{[pfSelRandMin not synced]}}
\providecommand{\pfSelRandomCod}{\textbf{[pfSelRandomCod not synced]}}
\providecommand{\pfSelRandomDis}{\textbf{[pfSelRandomDis not synced]}}
\providecommand{\pfSelRandomLl}{\textbf{[pfSelRandomLl not synced]}}
\providecommand{\pfSelRandomSod}{\textbf{[pfSelRandomSod not synced]}}

\providecommand{\pfTrnKCod}{\textbf{[pfTrnKCod not synced]}}
\providecommand{\pfTrnKDis}{\textbf{[pfTrnKDis not synced]}}

\providecommand{\pfTrnKSod}{\textbf{[pfTrnKSod not synced]}}

\providecommand{\pfTrnNCod}{\textbf{[pfTrnNCod not synced]}}
\providecommand{\pfTrnNDis}{\textbf{[pfTrnNDis not synced]}}

\providecommand{\pfTrnNSod}{\textbf{[pfTrnNSod not synced]}}

\providecommand{\pfTrnOracleCod}{\textbf{[pfTrnOracleCod not synced]}}
\providecommand{\pfTrnOracleDis}{\textbf{[pfTrnOracleDis not synced]}}

\providecommand{\pfTrnOracleSod}{\textbf{[pfTrnOracleSod not synced]}}

\providecommand{\pfTrnRandomCod}{\textbf{[pfTrnRandomCod not synced]}}
\providecommand{\pfTrnRandomDis}{\textbf{[pfTrnRandomDis not synced]}}

\providecommand{\pfTrnRandomSod}{\textbf{[pfTrnRandomSod not synced]}}

\providecommand{\xaPairACod}{\textbf{[xaPairACod not synced]}}
\providecommand{\xaPairADis}{\textbf{[xaPairADis not synced]}}
\providecommand{\xaPairADuts}{\textbf{[xaPairADuts not synced]}}

\providecommand{\xaPairCCod}{\textbf{[xaPairCCod not synced]}}
\providecommand{\xaPairCDis}{\textbf{[xaPairCDis not synced]}}
\providecommand{\xaPairCDuts}{\textbf{[xaPairCDuts not synced]}}

\providecommand{\xmFrameMax}{\textbf{[xmFrameMax not synced]}}
\providecommand{\xmFrameMin}{\textbf{[xmFrameMin not synced]}}

\title{Can Frozen Hyperspherical Features\\Guide the Selection of Pseudo Masks?}

\iclrfinalcopy
\author{%
\makebox[\dimexpr\textwidth-2\tabcolsep][c]{\bf Xinge Guo$^{1}$, Fengyang Xiao$^{1}$, Dingming Zhang$^{1}$, Yuhan Chen$^{1}$, Rihan Zhang$^{1}$}\\
\multicolumn{1}{c}{\bf Xingjian Li$^{2}$, Tianyang Wang$^{3}$, Chunming He$^{1,\dagger}$, Sina Farsiu$^{1}$}\\[5pt]
\multicolumn{1}{c}{\normalfont\normalsize $^{1}$Duke University}\\
\multicolumn{1}{c}{\normalfont\normalsize $^{2}$Carnegie Mellon University}\\
\multicolumn{1}{c}{\normalfont\normalsize $^{3}$University of Alabama at Birmingham}}

\newcommand\blfootnote[1]{%
  \begingroup\renewcommand\thefootnote{}\footnote{#1}\addtocounter{footnote}{-1}\endgroup}

\begin{document}
\maketitle
\lhead{\small Preprint. Under review.}
\blfootnote{$\dagger$ Corresponding Author}
\setcounter{tocdepth}{2}
\addtocontents{toc}{\protect\setcounter{tocdepth}{-1}}

\providecommand{\rot}[1]{\rotatebox{90}{\small #1}}

\definecolor{pipeblue}{RGB}{86,180,233}
\definecolor{pipeorange}{RGB}{230,159,0}
\definecolor{pipegreen}{RGB}{0,158,115}
\definecolor{pipepurple}{RGB}{204,121,167}
\newcommand{\pipebox}[2]{%
  \begingroup\setlength{\fboxsep}{2pt}%
  \fcolorbox{#1!75!black}{#1!12!white}{%
    \parbox[c][0.72cm][c]{0.19\linewidth}{\centering\scriptsize\bfseries #2}}%
  \endgroup}
\newcommand{\pipearrow}{\hspace{1.5pt}$\longrightarrow$\hspace{1.5pt}}

\begin{abstract}
\emergencystretch=1em\relax
Foundation segmenters such as SAM return several plausible masks for an unlabeled image, and a
student trained on the wrong one inherits its errors. Choosing among them means querying a second
large model or fitting a quality head to annotated masks. We show that a candidate can be judged by
what it does to a frozen self-supervised backbone's features. Normalized DINOv2 patch features lie on a
hypersphere, and a candidate mask splits that sphere in two. Based on this reading, we introduce
SPHERETRUST, which scores each candidate by three properties of the split, the angular contrast
between the two sides, the coverage of the foreground's appearance modes, and contact with the image
frame, one for each of three common ways a mask fails, and ranks a pool in $0.55$\,s per image from
the frozen features alone. On eight SAM and SAM3 candidate pools spanning camouflaged, salient, and
dichotomous segmentation and camouflage under low light, SPHERETRUST exceeds the strongest evaluated
external baseline on six pools by $1.7$ to $9.3$ percentage points in mean selected Dice. These
comparisons include published selection rules and explicitly labeled adaptations of DSS and
UCOD-MKD. On the two prompted camouflage pools, its mean selected Dice is within $0.1$ percentage
points of the candidate-derived DSS adaptation, with a lower catastrophic-error rate.
Which cue carries the signal depends on the candidate pool. The same sphere also supports training.
The leading candidates enter as a candidate set with their scores as priors, prototypes reorder
them, and a cross-fitted second round completes the labels, raising weighted $F$ by $4.5$, $2.3$,
and $5.5$ points over fixed-label training on the three MLLM anchor pools, with students competitive
with published unsupervised methods on nineteen test sets.
\end{abstract}

\section{Introduction}
\label{sec:intro}

Pseudo labels turn model predictions into training targets \citep{lee2013pseudo,sohn2020fixmatch},
and unsupervised segmentation learns from masks discovered automatically
\citep{wang2023cutler,wang2024unsam}. A student trained on such masks inherits their errors, an
omitted part, leaked background, or the wrong object. Foundation segmenters such as the Segment
Anything Model (SAM) return several plausible candidates per image
\citep{kirillov2023segment}, so the question shifts from producing a mask to choosing one, and the
generator judges its own output poorly. On a camouflage pool whose best candidates average
\fpOrc\ Dice~\citep{dice1945measures}, SAM predicted IoU picks masks averaging \fpSam\
(Appendix~Tab.~\ref{tab:diag}).

Existing selectors judge a candidate from the outside, by consensus among the generator's masks
\citep{shin2022selfmask,chen2026beyond}, by the contour alone, or with a second large model
\citep{yang2026dss}. We ask what the mask does to the image's own features.

\begin{figure}[!t]
\centering
\includegraphics[width=\linewidth]{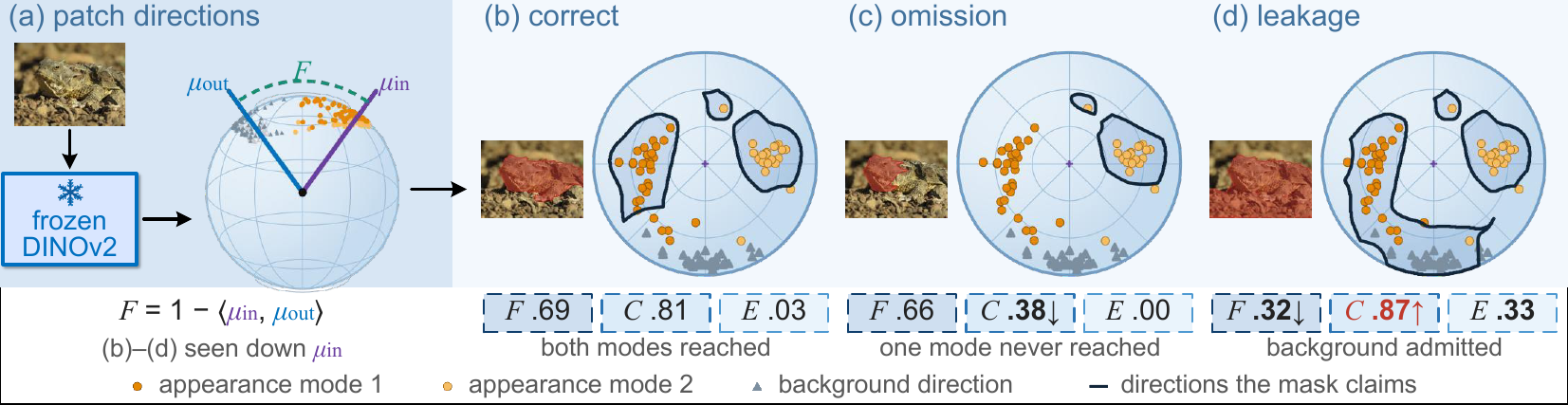}
\caption{\textbf{A mask as a partition of the sphere.} Frozen DINOv2 patch directions under a
correct mask, an omission, and leakage. The three measures respond differently to the two errors.}
\label{fig:mechanism}
\end{figure}

Self-supervised vision transformers place their normalized patch features on \mbox{$\Sph^{d-1}$}
\citep{caron2021emerging,oquab2024dinov2}, and LOST~\citep{simeoni2021lost} and
TokenCut~\citep{wang2022tokencut} search this geometry for an object. We are handed candidates and
ask which one the geometry already agrees with. A binary mask is a bipartition of the image's patch
directions, so a candidate's quality becomes a property of that partition. A good mask separates two
directional populations and reaches every
appearance mode of the foreground, a fragmented selection pulls the two mean
directions together, an omission leaves a mode uncovered, and contact with the frame marks the
background side (Fig.~\ref{fig:mechanism}). Angular contrast measures how sharply a candidate cuts
the directions, and a coherent distractor cuts just as sharply, so the target itself comes from the
pool's prompts (\S\ref{sec:e1}). SPHERETRUST reads the three properties as angular contrast
$\Fcon$, spherical coverage $\Csph$, and frame contact $\Ebd$, standardized within the image
and summed with equal weights. Contrast is invariant under complementation, so coverage and frame
contact tell the two sides apart, with a foreground prototype read off the patches most candidates
share and a background prototype off the border ring. A multimodal anchor enters only where two of
the three pools are built (\S\ref{sec:e2}).

This reading passes three tests (\S\ref{sec:e1}). On six of the eight pools, SPHERETRUST exceeds
the strongest evaluated external comparator by $1.7$ to $9.3$ percentage points in selected Dice;
on the two prompted camouflage pools, it is within $0.1$ percentage points of the candidate-derived
DSS adaptation, with a lower catastrophic-error rate. The comparisons include published selection
rules and explicitly labeled adaptations, and SPHERETRUST ranks candidates in $0.55$\,s per image.
For the tested median-margin statistic, normalized feature directions rank mask quality better than
raw features or feature norms, and angular contrast from SAM's own encoder yields less than half the
selected Dice. The pool's prompts provide target information while the score ranks candidate extent
and quality.

Selecting a pseudo mask and using it for training are two decisions. Training on a fixed top
ranked mask discards the other candidates as supervision. We
instead retain a few leading, distinct masks and use their geometric scores as priors for target
selection during training, so the student changes its target where a better fit outweighs the score
gap (\S\ref{sec:m2}).

Our scope is the DIS formulation \citep{qin2022dis}, a binary foreground/background mask, on
camouflaged, salient, and dichotomous segmentation and on camouflage under synthetic low light, at
the Vision Transformer (ViT) patch scale \citep{dosovitskiy2021image}. Both regimes are usually
studied with pixel labels from the target domain \citep{song2024basam,he2025run} or with
illumination restored first \citep{he2023retidiff,he2026unfoldir}. We make three contributions.
\begin{enumerate}
\item A new reading of pseudo mask quality. A candidate is judged by the partition it induces on the
hypersphere of a frozen self-supervised backbone, through three measures answering to three common
failures, wrong object, omission, and leakage, combined within each image.
\item A label-free selector built on that reading. Three standardized measures summed with equal
weights rank a pool's candidates from frozen features in $0.55$\,s per image. For the tested
median-margin statistic, normalized feature directions rank mask quality better than raw features
or feature norms (\S\ref{sec:e1}, Appendix~\S\ref{app:design}).
\item A training stage fed by the same sphere. The leading candidates enter training as a candidate
set with their scores as priors, prototypes on the same sphere reorder them, and a cross-fitted
second round completes the labels. The stage raises the student of each of three ranking rules by
$2.6$ to $10.7$ points of weighted $F$ on camouflaged, salient, and dichotomous data, showing that
the training stage improves on fixed-label training across all three tested ranking rules
(\S\ref{sec:e2}, Appendix~\S\ref{app:armcmp}).
\end{enumerate}

\section{Related Work}
\label{sec:related}

\PAR{Pseudo mask selection and quality estimation.}
Self training converts model predictions into targets \citep{lee2013pseudo} and keeps the confident
ones \citep{sohn2020fixmatch} or those consistent across views
\citep{tarvainen2017mean,berthelot2019mixmatch}, although \citet{guo2017calibration} show that
confidence can be miscalibrated. SAM exposes predicted IoU \citep{kirillov2023segment},
SAMRefiner ranks candidates by it \citep{lin2025samrefiner}, SAQ scores the contour alone
\citep{lin2024saq}, and learned quality models require labels or a reference set
\citep{huang2019maskscoring,robinson2018realtime,valindria2017reverse}. Two closely related
procedures are the feature-map and boundary score of DSS \citep{yang2026dss} and the mask-grading
stage of UCOD-MKD \citep{chen2026beyond}. DSS scores a mask against a similarity map built from a
separately discovered region, and UCOD-MKD selects by generator confidence and assigns a quality
tier using agreement among candidates and, where applicable, structural checks on the selected
mask. SPHERETRUST instead scores the bipartition of patch directions that the candidate itself
induces. We compare explicitly labeled adaptations of both on identical candidate pools
(\S\ref{sec:e1}, Appendix~\S\ref{app:closest}).

\begin{figure}[!t]
\centering
\includegraphics[width=\linewidth]{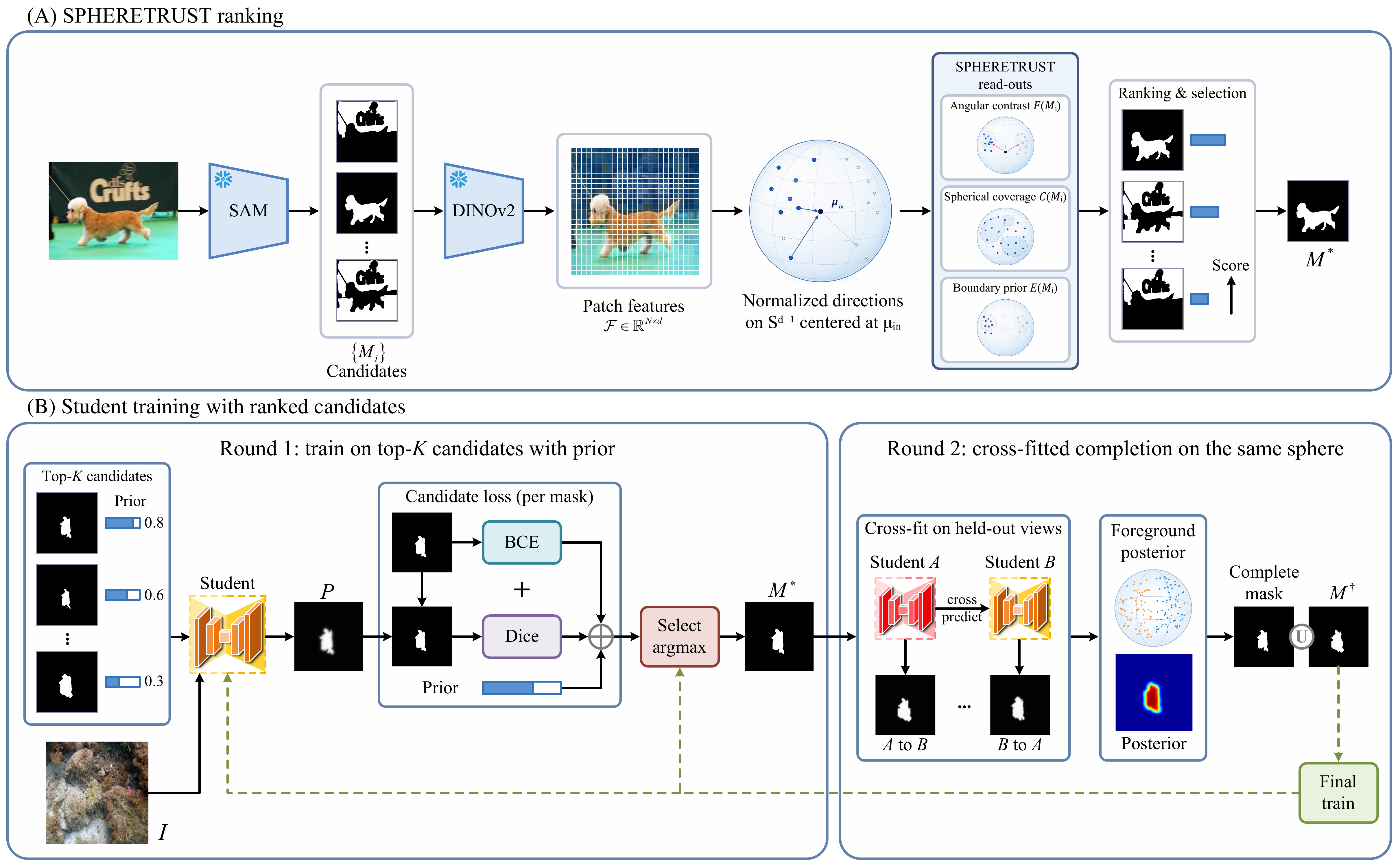}
\caption{\textbf{SPHERETRUST overview.} (A) Frozen DINOv2 directions rank candidates by angular
contrast, spherical coverage, and frame contact, where $\mathcal{F}$ is the patch feature matrix,
distinct from $\Fcon$ of Eq.~(\ref{eq:F}). (B) A foreground posterior on the same sphere joins the
ranking, the student chooses among the leading candidates as it learns, and a second round adds a
completed mask from a cross fitted prediction (\S\ref{sec:m2}). The \emph{select argmax} box in (B)
is Eq.~(\ref{eq:cs}), $\arg\max_k[\pi_{ik}-\ell_{ik}/T]$.}
\label{fig:pipeline}
\end{figure}

\PAR{Unsupervised object and camouflage segmentation.}
Surveyed for camouflage by \citet{xiao2024survey}, this literature spans object discovery, pseudo
mask refinement, knowledge transfer, proposal agreement, and priors from foundation models
\citep{wang2023cutler,wang2024unsam,yuan2024a2sv3,shin2022selfmask,zhou2023a2sv2,yan2025ucoddpl},
and UCOD-MKD \citep{chen2026beyond} distills from an MLLM and a SAM teacher. Camouflage frameworks
that build their own masks, by retrieval \citep{du2025rise}, attention shifting
\citep{shou2025sdalsnet}, or spectral refinement \citep{liu2026dualucod},
are \emph{generators} rather than selectors, so their published numbers
stand beside our students in Tab.~\ref{tab:t2} rather than in Tab.~\ref{tab:t1}.

\PAR{Candidate set supervision.}
Partial label learning trains on examples carrying a set of candidate labels and resolves the set
during training by the model's own fit \citep{lv2020proden,wang2022pico}, and dynamic pseudo label
schedules do the same for segmentation \citep{yan2025ucoddpl}. Eq.~(\ref{eq:cs}) is the spatial
dense form, with the generator's candidates as the label set and the score of \S\ref{sec:m1} as
the prior.

\section{Method}
\label{sec:method}

\PAR{Setup and notation.}
A frozen generator returns $K_i$ binary candidates
$\mathcal{M}_i=\{M_{ij}\}_{j=1}^{K_i}$ for image $i$, each held at image
resolution for the segmentation loss and on the binary feature grid for every sum over patches. A
frozen DINOv2 ViT-B/14 maps
each $350\times350$ image to a $G\times G$ grid of patch features
$f_p\in\R^{768}$ ($G=25$), normalized to
$\fhat_p=f_p/\lVert f_p\rVert_2\in\Sph^{767}$. For
any patch set $S$,
$\mu(S)=\sum_{p\in S}\fhat_p/(\lVert\sum_{p\in S}\fhat_p\rVert_2+\varepsilon_\mu)$, and
$\langle\cdot,\cdot\rangle$ is the inner product. The anchor region $\Omega$ is the bounding box of
the candidate union in an unprompted pool, or the union of the expanded
prompt boxes in a prompted pool, and $M_\Omega=M\cap\Omega$.

\subsection{Ranking by the Induced Partition}
\label{sec:m1}

\PAR{Angular contrast.}
Writing $\bar{M}$ for the grid minus $M$, we define \emph{angular contrast} as how far apart the
partition drives the two mean directions,
\begin{equation}
\label{eq:F}
\Fcon(M)=1-\big\langle\mu(M),\,\mu(\bar{M})\big\rangle\in[0,2].
\end{equation}
$\Fcon$ is small when $M$ splits one directional population and large when it separates two.

\PAR{The spherical dictionary.}
Coverage needs an estimate of which directions in $\Omega$ resemble foreground, built once per image
from the frozen features and candidates. A foreground prototype $\phi_{\mathrm{fg}}=\mu(\Omega_{\mathrm{core}})$ is taken over the \emph{core}
$\Omega_{\mathrm{core}}\subseteq\Omega$, the patches covered by at least half of that image's candidates, and a
background prototype $\phi_{\mathrm{bg}}=\mu(\mathcal{B})$ over the border ring $\mathcal{B}$, two
patches wide, with fallbacks that keep the core populated (Appendix~\S\ref{app:impl}). Euclidean
$k$-means on the unit
directions of the candidate union inside $\Omega$, with the fitted centroids projected back to the
sphere, the usual substitute for spherical $k$-means \citep{dhillon2001spherical}, gives at most $J=4$ centroids
$\mu_1,\dots,\mu_J$, the image's \emph{appearance modes}, of which the
foreground modes are those closer to $\phi_{\mathrm{fg}}$ than to $\phi_{\mathrm{bg}}$,
$\mathcal{K}_f=\{j:\langle\mu_j,\phi_{\mathrm{fg}}\rangle>\langle\mu_j,\phi_{\mathrm{bg}}\rangle\}$,
falling back to the best aligned mode when the set is empty. A patch $p\in\Omega$ goes to its
nearest foreground mode, $a(p)=\argmax_{j\in\mathcal{K}_f}\langle\fhat_p,\mu_j\rangle$, and counts as
foreground when that mode explains it better than the background prototype does,
$s(p)=\mathbf{1}[\max_{j\in\mathcal{K}_f}\langle\fhat_p,\mu_j\rangle>\langle\fhat_p,\phi_{\mathrm{bg}}\rangle]$.

\PAR{Spherical coverage.}
Let $\Omega_j=\{p\in\Omega: s(p)=1,\,a(p)=j\}$ be the foreground patches that mode $j$ owns and
$r_j(M)=|M\cap\Omega_j|/(|\Omega_j|+\varepsilon_C)$, with
$\varepsilon_C=10^{-8}$ ($\Omega_j\subseteq\Omega$, so restricting $M$ to $\Omega$ changes
nothing). \emph{Spherical coverage} is the average of this recall over
all foreground modes,
\begin{equation}
\label{eq:C}
\Csph(M)=\frac{1}{|\mathcal{K}_f|}\sum_{j\in\mathcal{K}_f}r_j(M).
\end{equation}
An empty $\Omega_j$ scores zero and stays in the average. $\Csph$ is
nondecreasing as patches are added, so it detects omissions but may reward leakage
(Fig.~\ref{fig:mechanism}).

\PAR{One spatial prior.}
With $\partial I$ the outermost pixel ring of the mask at the $350\times350$ working resolution of
the backbone, the border term is the fraction of that ring the mask occupies,
\begin{equation}
\label{eq:E}
\Ebd(M)=\frac{|\partial I\cap M|}{|\partial I|}\in[0,1],
\end{equation}
which the rule subtracts. It is the boundary prior of unsupervised saliency
\citep{zhu2014saliency}, and the candidate filter keeps its range to $[0,\tfrac{1}{2})$.

\begin{figure}[!t]
\centering
\includegraphics[width=\linewidth]{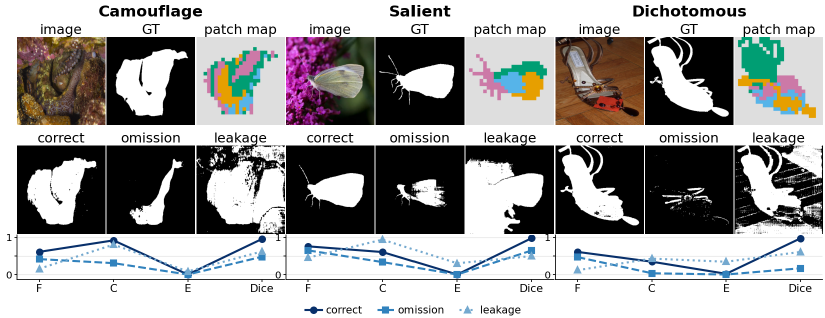}
\caption{\textbf{The three cues on real candidates.} For one image per regime, the patch map coloured by
the nearest foreground appearance mode and three candidates with their $\Fcon$, $\Csph$,
$\Ebd$, and Dice.}
\label{fig:cuesreal}
\end{figure}

\PAR{Combined selection score.}
The three terms have different ranges, so we standardize each within one image's
candidate set. For statistic $x$, we compute
$z_i(x_{ij})=(x_{ij}-\bar{x}_i)/(\sigma_i(x)+\varepsilon_z)$ if
$\sigma_i(x)>\varepsilon_z$ and $0$ otherwise ($\varepsilon_z=10^{-6}$, $\sigma_i$ the population
standard deviation), and sum the guarded terms with equal weights,
\begin{equation}
\label{eq:R}
\STscore(M_{ij})=z_i\!\big(\Fcon(M_{ij})\big)+z_i\!\big(\Csph(M_{ij})\big)
-z_i\!\big(\Ebd(M_{ij})\big),
\qquad
\hat{\jmath}_i=\argmax_j \STscore(M_{ij}).
\end{equation}
We use this untrained combination in every evaluation, and exact ties fall back to generator
confidence. Two of the three
terms read the mask on the $25\times25$ patch grid, so masks differing inside a patch score alike
(\S\ref{sec:disc}). Fig.~\ref{fig:cuesreal} shows the three terms on real candidates.

\subsection{From Geometric Ranking to Candidate Set Training}
\label{sec:m2}

Eq.~(\ref{eq:R}) selects one mask per image, and training can revisit that choice as the student
improves. We retain the leading, distinct candidates and use their geometric scores as priors for
the active supervision, which after a short fixed label phase balances them against the
student's current loss.

\PAR{Sphere term.}
Eq.~(\ref{eq:R}) uses one image at a time, while the selected masks of the whole pool fix which
directions foreground patches take on the same sphere. Von Mises--Fisher prototypes
\citep{mardia2000directional,banerjee2005clustering} fitted by spherical
$k$-means on the block $12$ directions of one half of the training images, labeled by their selected
masks (Appendix~\S\ref{app:impl}), give through the log
likelihood ratio of the two mixtures a foreground posterior $G_i(p)$ for every patch of the other
half, and vice versa. The sphere term
$\Gamma(M)$ is the mean of $G_i$ inside $M$ minus the mean outside, and the training order is
$\STscore(M_{ij})+z_i(\Gamma(M_{ij}))$.

\PAR{Candidate set training.}
For image $i$ let $M_{i1},\dots,M_{iK}$ be its candidates in decreasing training order after greedy
removal of any candidate whose IoU with a kept one reaches $0.8$, with $K=5$, and let $\pi_{ik}$ be
the training score of each. Here $k$ indexes the retained set and $j$ the pool, as in
Eq.~(\ref{eq:R}). With $\hat{y}_i$ the student's soft prediction and $\ell_{ik}$ the
BCE$+$Dice loss of $\hat{y}_i$ against $M_{ik}$, the image is supervised by the candidate that
balances its geometric score against the student's current fit,
\begin{equation}
\label{eq:cs}
k_i^\star=\argmax_k\big[\pi_{ik}-\ell_{ik}/T\big],\qquad
\mathcal{L}=\textstyle\sum_i \ell_{ik_i^\star},
\end{equation}
with $T=0.1$ and the choice detached from the gradient. During the first five of $25$
epochs $k_i^\star=1$, which is training on one fixed label, and afterwards an image moves to a lower
ranked candidate only where the student's fit outweighs the score gap, the dense form of partial
label learning \citep{lv2020proden,wang2022pico}.

\PAR{Second round.}
Round two obtains a label free signal by \emph{cross fitting}. The
training images are split into the same two halves and a round one student trained on each predicts
the other. Each image is predicted by a student trained on images in the other fold. However,
the prototype-based ordering of that student's training candidates uses pseudo-label information from
the predicted fold. These are out-of-fold predictions with shared prototype information. Removing the
prototype term eliminates this route but does not establish that the route has no effect in the full
method (Tab.~\ref{tab:t3b}b). Writing $P_i$ for that prediction,
candidates are rescored by
\begin{equation}
\label{eq:S}
S(M_{ij})=\STscore(M_{ij})+z_i\big(\Gamma(M_{ij})\big)+z_i\big(\mathrm{IoU}(M_{ij},P_i)\big),
\end{equation}
with the soft intersection over union against $P_i$. The \emph{completed mask}
$U_i=M_i^{S}\cup\{P_i\ge\tau\}$, the leading candidate $M_i^S$ under $S$ joined with the confident
region of the prediction, is placed first in a new candidate set, followed by the four leading
distinct candidates under $S$. $U_i$ inherits the prior of $M_i^S$, which always stays available
(Appendix~\S\ref{app:impl}). The final student is trained on this set by Eq.~(\ref{eq:cs}) and
decides per image between the completed mask and the plain candidates (Appendix~\S\ref{app:cs}).
Its training order and
$\tau=0.9$ were chosen on validation splits held out of the training pools (\S\ref{sec:exp}), and
its fold students, prototypes and candidate sets come from seed $0$ and are shared by the three
final seeds.

\section{Experiments}
\label{sec:exp}

\PAR{Datasets and protocol.}
We evaluate mask selection on shared candidate pools and the downstream students over
\nDatasets\ sets in four regimes, reporting
COD10K~\citep{fan2020camouflaged},
DUTS-TE~\citep{wang2017duts}, DIS-VD~\citep{qin2022dis}, and LL-COD, synthetic
degradation of the COD10K test images scored with the camouflage student (Appendix~\S\ref{app:regime2}).
P1 is the unprompted pool capped at twelve candidates, and P2 the prompted training pool, with the
candidates of every anchor box merged (Appendix~\S\ref{app:impl}).
Selected Dice is the quality of the chosen mask, Top-1 the fraction of images on which the selector
picks the highest Dice candidate \citep{lin2025samrefiner}, and catastrophic error is
$\Pr[D_{\mathrm{sel}}<0.2]$, all conditional on the eligibility filter of
Appendix~\S\ref{app:floors}. One evaluator computes $S_\alpha$, $F_\beta^w$,
$E_\phi$, and MAE \citep{fan2017structure,margolin2014evaluate,fan2018enhanced} for every downstream
method with released predictions, and transcribed rows are compared only under the same metric
definition (Appendix~\S\ref{app:t2full}). All students share the
architecture, schedule, and three seeds, and the two open settings of the training stage were
chosen on validation splits by one rule for all regimes, with the test sets scored after that
choice (Appendix~Tab.~\ref{tab:config}).

\PAR{Label use.}
The selector uses no ground-truth annotations. The downstream training configuration was selected
using $1{,}085$ annotated validation images. The label-free claim therefore applies to the selector,
and downstream model development uses validation supervision (Appendix~\S\ref{app:impl}).

\subsection{Selection on a Fixed Candidate Pool}
\label{sec:e1}

\begin{table}[!htbp]
\caption{\textbf{Unsupervised selection on prompted pools}, the camouflage training pool SAM3 builds
alone and the three training pools prompted from the MLLM anchor with SAM
(Appendix~Tab.~\ref{tab:floors}). \best{Red}, \second{blue} and \third{grey} mark the first, second and third value in
each column among the evaluated external comparators, the published selection rules and the
labeled adaptations, our row is shaded, and the rows below the rule, the size control and the
oracle, stay unmarked.}
\label{tab:t1}
\centering\scriptsize\setlength{\tabcolsep}{3.6pt}
\begin{tabularx}{\linewidth}{@{}lYYYY@{}}
\toprule
Selector & \textbf{SAM3 alone} & \multicolumn{3}{c@{}}{\textbf{MLLM anchor $+$ SAM}} \\
\multicolumn{1}{@{}l}{} & \footnotesize\itshape COD & \footnotesize\itshape COD & \footnotesize\itshape SOD & \footnotesize\itshape DIS \\
\midrule
\multicolumn{5}{@{}l@{}}{\emph{each entry:}~~sel-Dice$\uparrow$ / Top-1$\uparrow$ / Dice$<$.2$\downarrow$} \\
\addlinespace[1pt]
Random & .657/.096/.193 & .678/.057/.170 & .771/.045/.111 & .576/.047/.204 \\
SelfMask vote~\citep{shin2022selfmask} & \third{.821}/.124/\second{.062} & \third{.778}/.060/\second{.085} & .887/.034/.024 & .701/.060/\third{.086} \\
SAQ~\citep{lin2024saq} & .576/.109/.275 & .676/.098/.184 & .771/.163/.125 & .606/.117/.187 \\
Generator confidence~\citep{carion2025sam3} & .802/.145/.082 & .728/.057/.135 & .874/.033/.040 & .636/.060/.179 \\
Generator stability~\citep{carion2025sam3} & .688/.138/.187 & .733/.071/.135 & .876/.089/.038 & .620/.080/.196 \\
TokenCut~\citep{wang2022tokencut} & .730/\third{.215}/.121 & .707/\third{.177}/.130 & \third{.912}/\best{.284}/.025 & .688/\best{.309}/.142 \\
Generator confidence with UCOD-MKD-inspired structural filtering & .804/.137/.078 & .772/.070/\third{.088} & .889/.035/\second{.018} & \third{.715}/.075/\second{.071} \\
Adapted DSS scoring rule (candidate-derived map) & \best{.825}/\best{.252}/\third{.066} & \best{.788}/\second{.198}/.093 & \second{.912}/\third{.197}/\third{.019} & \second{.720}/\third{.171}/.095 \\
Adapted DSS scoring rule (anchor-box map) & .620/.132/.227 & .549/.099/.305 & .839/.181/.080 & .606/.148/.199 \\
\rowcolor{ourfill}
\textcolor{rkbest}{\textbf{SPHERETRUST}} & \second{.824}/\second{.251}/\best{.049} & \second{.787}/\best{.199}/\best{.073} & \best{.929}/\second{.270}/\best{.010} & \best{.771}/\second{.262}/\best{.050} \\
\midrule
\emph{size control:} $z(\Fcon)+z(\log\text{area})-z(\Ebd)$ & .822/.280/.053 & .763/.231/.089 & .909/.419/.024 & .755/.289/.062 \\
oracle & .893/1.000/.018 & .852/1.000/.040 & .949/1.000/.006 & .839/1.000/.016 \\
\bottomrule
\end{tabularx}

\end{table}

\begin{figure}[!b]
\centering
\includegraphics[width=\linewidth]{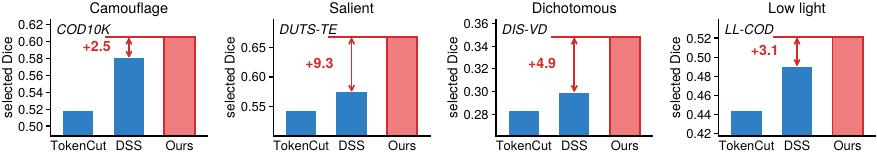}
\caption{\textbf{Selection where it is hardest.} Selected Dice on the four unprompted SAM pools of
Appendix~Tab.~\ref{tab:t1unp}, where a selector must first find the target among decoys, for the two
strongest evaluated external comparators, TokenCut and the candidate-derived adaptation of DSS, and
SPHERETRUST, margins in points.}
\label{fig:pools}
\vspace{5pt}
\begin{minipage}[c]{0.355\linewidth}\centering
\includegraphics[width=\linewidth]{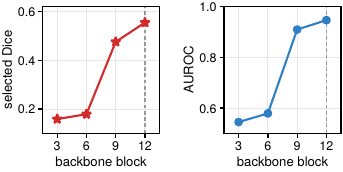}
\end{minipage}\hfill
\begin{minipage}[c]{0.625\linewidth}\centering
\includegraphics[width=\linewidth]{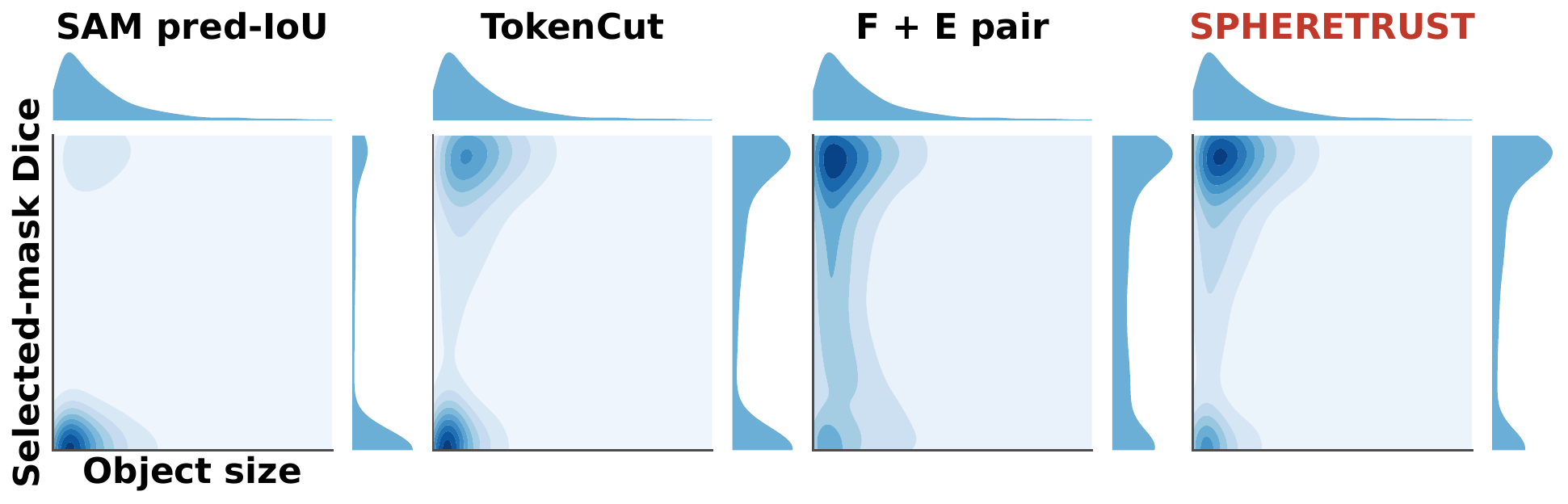}
\end{minipage}
\caption{\textbf{Depth of the signal, and where the selected masks move.} Left, angular contrast
alone from each probed block on the unprompted COD10K test pool. Right, target size against selected
Dice on the P1 camouflage training subset for four selectors. Mass leaves the lower left, small
targets with a wrong or partial mask, and the full score selects $.038$ higher Dice than the
$\Fcon+\Ebd$ pair (Appendix~\S\ref{app:terms}).}
\label{fig:density}
\end{figure}

Table~\ref{tab:t1} compares every selector on the same candidates of the SAM3 pool and of the three
MLLM anchor pools (Appendix~\S\ref{app:floors}). Two rows adapt closely related methods to
identical candidate pools. The \textbf{DSS adaptation} uses Eq.~(7) of \citet{yang2026dss},
substituting a similarity map derived from the candidate's own foreground and our border term for
the original discovery map and border measure (Appendix~\S\ref{app:closest}). A second adaptation
reads the map from the anchor box instead and is weaker on all four pools, so the margins below use
the stronger of the two. The
\textbf{UCOD-MKD-inspired adaptation} applies structural filtering before selecting by generator
confidence. SPHERETRUST has the highest selected Dice on the salient and dichotomous pools, by $1.7$
and $5.2$ points over adapted DSS, the strongest evaluated external comparator on both. On the two
camouflage pools, its selected Dice is within $0.1$ percentage points of that adaptation, $.7873$
against $.7881$ and $.8241$ against $.8245$. Among the evaluated external comparators and
SPHERETRUST, it has the lowest catastrophic-error rate on all four prompted pools and the
second-highest Top-1 on three of four. Generator confidence trails it by $2.2$ points on the SAM3
pool and $5.5$ to $13.6$ on the SAM pools (all margins use unrounded values).
Fig.~\ref{fig:selprompted} shows the pick of each rule on two images per pool. On the four
unprompted SAM pools of Appendix~\S\ref{app:unprompted}, the margin over the strongest evaluated
external comparator is $2.5$ to $9.3$ points (Fig.~\ref{fig:pools}), and Fig.~\ref{fig:density}
shows where the selected masks move.

\begin{table}[tb]
\caption{\textbf{Ranking and training ablations on the MLLM anchor pools.} (a) Selected Dice for
all seven combinations of the three ranking terms, with random and oracle selection as references.
(b) $F^w_\beta$ of the student, mean over the regime's test sets and three seeds, with candidate
set training (set), the sphere term in the round one order ($\Gamma_1$), the second round (R2), the
completed mask ($U$), cross fitting (XF), and the sphere term in the second round score
($\Gamma_2$) on or off. Marks as in Tab.~\ref{tab:t1}, reference rows stay unmarked, and
Appendix~Tab.~\ref{tab:csabl} adds the seed spreads.}
\label{tab:t3a}\label{tab:t3b}
\centering
{\scriptsize
\begin{minipage}[t]{0.40\linewidth}\vspace{0pt}\centering
\textbf{(a) Ranking}\par\vspace{2pt}
\setlength{\tabcolsep}{4.5pt}\renewcommand{\arraystretch}{0.948}
\begin{tabular}{@{}ccc>{\raggedleft\arraybackslash}p{0.95cm}>{\raggedleft\arraybackslash}p{0.95cm}>{\raggedleft\arraybackslash}p{0.95cm}@{}}
\toprule
$\Fcon$ & $\Csph$ & $\Ebd$ & COD & SOD & DIS \\
\midrule
\multicolumn{3}{@{}l}{random candidate} & \pfTrnRandomCod & \pfTrnRandomSod & \pfTrnRandomDis \\
\multicolumn{3}{@{}l}{oracle candidate} & \pfTrnOracleCod & \pfTrnOracleSod & \pfTrnOracleDis \\
\midrule
\yes & \no & \no & \third{.7663} & .8789 & .6351 \\
\no & \yes & \no & .6587 & .8894 & .6761 \\
\no & \no & \yes & .7490 & .8269 & .6108 \\
\yes & \yes & \no & .7546 & \second{.9280} & \second{.7356} \\
\yes & \no & \yes & \second{.7789} & .8918 & .6741 \\
\no & \yes & \yes & .7112 & \third{.9049} & \third{.7339} \\
\midrule
\rowcolor{ourfill}
\yes & \yes & \yes & \best{.7873} & \best{.9288} & \best{.7714} \\
\bottomrule
\end{tabular}

\end{minipage}\hfill
\begin{minipage}[t]{0.575\linewidth}\vspace{0pt}\centering
\textbf{(b) Training}\par\vspace{2pt}
\setlength{\tabcolsep}{3.7pt}\renewcommand{\arraystretch}{0.828}
\begin{tabular}{@{}cccccc>{\raggedleft\arraybackslash}p{0.95cm}>{\raggedleft\arraybackslash}p{0.95cm}>{\raggedleft\arraybackslash}p{0.95cm}@{}}
\toprule
set & $\Gamma_1$ & R2 & $U$ & XF & $\Gamma_2$ & COD & SOD & DIS \\ \midrule
\multicolumn{6}{@{}l}{oracle candidate, fixed} & .8048 & .8311 & .6745 \\ \midrule
\no&\no&\no&\no&\no&\no & .7513 & .8148 & .6048 \\
\no&\yes&\no&\no&\no&\no & .7825 & .8211 & .6261 \\
\yes&\yes&\no&\no&\no&\no & .7930 & .8232 & .6384 \\
\yes&\yes&\yes&\no&\yes&\yes & \third{.7941} & .8209 & .6434 \\
\yes&\yes&\yes&\yes&\no&\yes & \second{.7953} & .8280 & .6475 \\
\no&\yes&\yes&\yes&\yes&\yes & .7866 & .8330 & .6570 \\
\yes&\no&\yes&\yes&\yes&\yes & .7932 & \third{.8368} & \second{.6579} \\
\yes&\no&\yes&\yes&\yes&\no & .7896 & \best{.8377} & \third{.6573} \\
\midrule
\rowcolor{ourfill}
\yes&\yes&\yes&\yes&\yes&\yes & \best{.7961} & \second{.8376} & \best{.6594} \\
\bottomrule
\end{tabular}

\end{minipage}}
\end{table}

\begin{figure}[!t]
\centering
\includegraphics[width=\linewidth]{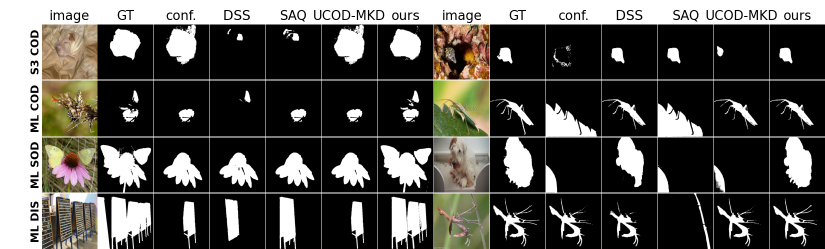}
\caption{\textbf{Selected masks on the prompted pools.} For two images of each pool, the SAM3 pool
(S3) and the three MLLM anchor pools (ML), the ground truth and the candidate chosen by generator
confidence (conf.), the candidate-derived adaptation of DSS (DSS), SAQ, UCOD-MKD-inspired filtering,
and SPHERETRUST (ours) from one candidate list.}
\label{fig:selprompted}
\end{figure}

\PAR{Contribution of each term.}
Table~\ref{tab:t3a}(a) tests all seven term combinations, and the full score has the highest
selected Dice on all three pools, by $.008$ and $.036$ over the best pair on COD and
DIS and by $.001$ on SOD, and equal
weights match or exceed fitted linear predictors (Appendix~\S\ref{app:learned}). Coverage correlates
with area, so the size control of Tab.~\ref{tab:t1} replaces $z(\Csph)$ by $z(\log\text{area})$, and
coverage selects better on every prompted pool and on the decoder pool (Appendix~\S\ref{app:terms}).

\PAR{Direction, backbone, and which object is the target.}
For the tested median-margin statistic (Appendix~\S\ref{app:design}), normalized directions give
AUROC $.934$, against $.899$ for raw features and $.752$ for the norm alone, so normalization helps
that statistic while magnitude still carries some signal. Angular contrast gives AUROC $.946$
whether patches are normalized before averaging or raw features are averaged; this equality is
empirical and does not imply mathematical invariance of the raw-feature version. From SAM's own encoder $\Fcon$ gives $\bbSdice$ selected Dice against
$\bbBdice$ from DINOv2 ViT-B, and from the earlier blocks of that backbone far less
(Fig.~\ref{fig:density}, left). A
\emph{decoy} occurs in $91\%$ of unprompted COD training images, and the rule ranks candidates
overlapping the target above decoys at AUC $0.981$ (Appendix~\S\ref{app:unprompted}).

\subsection{Training the Student}
\label{sec:e2}

Section~\ref{sec:e1} tested the sphere as a selector. This section evaluates a second use of the
same frozen features in training, with ablations of the training components and a matched
comparison across selection rules. Selection and training use the frozen features and the
candidate masks alone (Fig.~\ref{fig:pipeline}).

\begin{table}[!t]
\caption{\textbf{Downstream comparison} on one benchmark per condition, marks as in
Tab.~\ref{tab:t1}. Rows marked $\dagger$ are transcribed, each camouflage method in its strongest
published configuration (Appendix~\S\ref{app:t2full}), other rows use our evaluator on released
predictions or weights, and the Selfment rows are its published result and its
released checkpoint under our evaluation. \emph{Selection only} trains the same student on the
fixed mask $\STscore$ selects. The rank marks cover the published methods and our full method,
leaving out that row and the last block, which prompts foundation models at test time.}
\label{tab:t2}
\centering
{\setlength{\tabcolsep}{1.0pt}\renewcommand{\arraystretch}{0.82}\scriptsize
\begin{minipage}[t]{0.50\linewidth}\vspace{0pt}
\begin{tabular}{@{}l>{\raggedleft\arraybackslash}p{0.55cm}>{\raggedleft\arraybackslash}p{0.55cm}>{\raggedleft\arraybackslash}p{0.55cm}>{\raggedleft\arraybackslash}p{0.70cm}@{}}
\toprule
Method & $S_\alpha$$\uparrow$ & $F^w_\beta$$\uparrow$ & $E_\phi$$\uparrow$ & MAE$\downarrow$ \\
\midrule
\multicolumn{5}{@{}l@{}}{\textbf{Camouflage}\quad\itshape COD10K, $n=2026$} \\
\addlinespace[1pt]
EReCu~\citep{jiang2026erecu}\smash{$^\dagger$} & .7221 & .5628 & .8185 & .0613 \\
FOUND~\citep{simeoni2023found}\smash{$^\dagger$} & .6783 & .5056 & .6475 & .0841 \\
TokenCut~\citep{wang2022tokencut}\smash{$^\dagger$} & .6638 & .4770 & .7539 & .1023 \\
UCOD-DPL (DINOv2)~\citep{yan2025ucoddpl}\smash{$^\dagger$} & .8340 & \second{.7630} & \second{.9160} & \second{.0310} \\
UCOD-MKD (PVTv2)~\citep{chen2026beyond}\smash{$^\dagger$} & .8350 & .7400 & \third{.9080} & \second{.0310} \\
RISE (SAM masks)~\citep{du2025rise}\smash{$^\dagger$} & .7900 & .6430 & .8540 & .0440 \\
DualUCOD (DINOv2)~\citep{liu2026dualucod}\smash{$^\dagger$} & .7610 & .6350 & .8560 & .0420 \\
Selfment, published~\citep{you2026selfment}\smash{$^\dagger$} & \best{.8730} & \third{.7540} & .9000 & .0330 \\
Selfment, our evaluation~\citep{you2026selfment} & .8253 & .5507 & .8439 & .0577 \\
A2S-v3~\citep{yuan2024a2sv3} & .6704 & .4762 & .7415 & .0960 \\
CSNet-crf~\citep{guan2025csnet} & .6238 & .3879 & .6259 & .0740 \\
\emph{selection only}, fixed label & .8253 & .7018 & .8710 & .0442 \\
\rowcolor{ourfill}
\textcolor{rkbest}{\textbf{SPHERETRUST}}, MLLM anchor & \third{.8428} & .7446 & .9007 & .0357 \\
\rowcolor{ourfill}
\textcolor{rkbest}{\textbf{SPHERETRUST}}, SAM3 pool & \second{.8613} & \best{.7733} & \best{.9279} & \best{.0239} \\
\midrule
\multicolumn{5}{@{}l@{}}{\textbf{Low light}\quad\itshape LL-COD, $n=2026$} \\
\addlinespace[1pt]
Selfment, our evaluation~\citep{you2026selfment} & \second{.8183} & \third{.5452} & \third{.8261} & \third{.0576} \\
A2S-v3~\citep{yuan2024a2sv3} & .6243 & .4031 & .7070 & .0991 \\
CSNet-crf~\citep{guan2025csnet} & .5626 & .2536 & .5054 & .0795 \\
\emph{selection only}, fixed label & .7811 & .6315 & .8333 & .0594 \\
\rowcolor{ourfill}
\textcolor{rkbest}{\textbf{SPHERETRUST}}, MLLM anchor & \third{.8001} & \second{.6791} & \second{.8625} & \second{.0513} \\
\rowcolor{ourfill}
\textcolor{rkbest}{\textbf{SPHERETRUST}}, SAM3 pool & \best{.8274} & \best{.7191} & \best{.9030} & \best{.0304} \\
\addlinespace[2.5pt]
\bottomrule
\end{tabular}

\end{minipage}\hfill
\begin{minipage}[t]{0.49\linewidth}\vspace{0pt}
\begin{tabular}{@{}l>{\raggedleft\arraybackslash}p{0.55cm}>{\raggedleft\arraybackslash}p{0.55cm}>{\raggedleft\arraybackslash}p{0.55cm}>{\raggedleft\arraybackslash}p{0.70cm}@{}}
\toprule
Method & $S_\alpha$$\uparrow$ & $F^w_\beta$$\uparrow$ & $E_\phi$$\uparrow$ & MAE$\downarrow$ \\
\midrule
\multicolumn{5}{@{}l@{}}{\textbf{Salient}\quad\itshape DUTS-TE, $n=5019$} \\
\addlinespace[1pt]
3SD~\citep{yasarla20243sd} & .8121 & .6370 & .7859 & .0869 \\
CutLER~\citep{wang2023cutler} & .6503 & .5149 & .6578 & .2180 \\
FOUND~\citep{simeoni2023found} & .8029 & .7479 & .8709 & .0579 \\
UMNet~\citep{wang2022umnet} & .8027 & .7039 & .8448 & .0667 \\
Selfment, our evaluation~\citep{you2026selfment} & .7977 & .5659 & .7737 & .1139 \\
A2S-v3~\citep{yuan2024a2sv3} & \third{.8477} & \third{.7930} & \third{.9040} & .0473 \\
CSNet-crf~\citep{guan2025csnet} & .7964 & .7275 & .8457 & .0562 \\
CSNet (full)~\citep{guan2025csnet} & \second{.8532} & \second{.8187} & \best{.9110} & \best{.0431} \\
TSD~\citep{zhou2023a2sv2} & .8424 & .7835 & .9019 & \third{.0468} \\
\emph{selection only}, fixed label & .8629 & .7993 & .8929 & .0470 \\
\rowcolor{ourfill}
\textcolor{rkbest}{\textbf{SPHERETRUST}}, MLLM anchor & \best{.8752} & \best{.8232} & \second{.9071} & \second{.0450} \\
\midrule
\multicolumn{5}{@{}l@{}}{\textbf{Dichotomous}\quad\itshape DIS-VD, $n=470$} \\
\addlinespace[1pt]
Selfment, our evaluation~\citep{you2026selfment} & \second{.7410} & \third{.4673} & .6920 & \third{.1469} \\
A2S-v3~\citep{yuan2024a2sv3} & \third{.6247} & \second{.4844} & \second{.7235} & .1545 \\
CSNet-crf~\citep{guan2025csnet} & .5940 & .4300 & \third{.7015} & \second{.1376} \\
\emph{selection only}, fixed label & .7333 & .6016 & .7849 & .1095 \\
\rowcolor{ourfill}
\textcolor{rkbest}{\textbf{SPHERETRUST}}, MLLM anchor & \best{.7599} & \best{.6516} & \best{.8233} & \best{.0968} \\
\addlinespace[2.5pt]
\multicolumn{5}{@{}l@{}}{\emph{COD10K, prompting SAM, SAM2, or MLLM at test time}} \\
EASE $+$ SAM~\citep{du2025ease}\smash{$^\dagger$} & .8560 & .8010 & .9140 & .0280 \\
EASE $+$ HQ-SAM~\citep{du2025ease}\smash{$^\dagger$} & .8660 & .8110 & .9180 & .0230 \\
DSS (SAM2, MLLM)~\citep{yang2026dss}\smash{$^\dagger$} & .8870 & .8490 & .9420 & .0220 \\
\bottomrule
\end{tabular}

\end{minipage}}
\end{table}

\begin{figure}[!t]
\centering
\begin{minipage}[t]{0.46\linewidth}\vspace{0pt}\centering
\includegraphics[width=\linewidth]{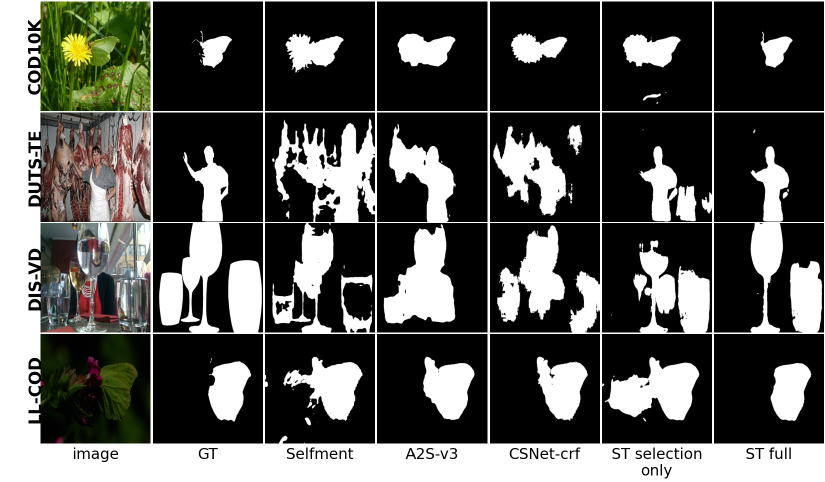}
\caption{\textbf{Downstream predictions} on the four conditions of Tab.~\ref{tab:t2}, one row per
condition, against three published methods and the selection only student.}
\label{fig:down}
\end{minipage}\hfill
\begin{minipage}[t]{0.52\linewidth}\vspace{0pt}\centering
\includegraphics[width=\linewidth]{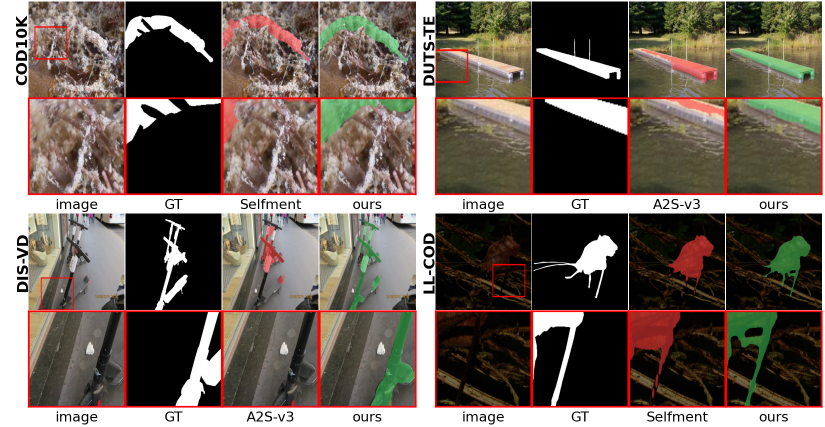}
\caption{\textbf{Where the student differs.} One hard case per condition against a strong
published method, with the boxed region magnified beneath.}
\label{fig:zoom}
\end{minipage}
\end{figure}

\PAR{Candidate set training against a fixed top ranked label.}
Table~\ref{tab:t3b}(b) separates how candidates are ranked from how they are used for
supervision.
Under the same round-one ordering $\STscore+z(\Gamma)$, candidate-set training improves weighted $F$
by $1.1$, $0.2$, and $1.2$ points on COD, SOD, and DIS relative to a fixed label selected by that
ordering. The complete procedure improves it by $4.5$, $2.3$, and $5.5$ points relative to the
original fixed label selected by $\STscore$, with gains of at least $4.3$, $1.7$, and $5.3$ points
on every seed.
The prototype term adds $3.1$, $0.6$, and $2.1$ points to fixed label training, but removing it from
the complete procedure changes the score by only $-0.65$, $+0.01$, and $-0.21$ points (Appendix
Tab.~\ref{tab:csabl}), so its incremental effect depends on the surrounding procedure.
On SOD the full method exceeds a student trained on the pool's oracle candidate ($.8376$ against
$.8311$), a fixed label student rather than an annotation ceiling, and closes $84\%$ and $78\%$ of
that gap on COD and DIS. Table~\ref{tab:t2} places the students among published methods, with a
selection only row so the training stage's margin can be read directly, and Figs.~\ref{fig:down}
and \ref{fig:zoom} show the predictions. The MLLM
anchor student leads $S_\alpha$ and $F^w_\beta$ on
DUTS-TE and every metric on DIS-VD, and on COD10K the SAM3 pool student is first in $F^w_\beta$,
$E_\phi$, and MAE among the methods above the rule of Tab.~\ref{tab:t2}, and second in $S_\alpha$
to the published Selfment row \citep{you2026selfment}, whose released checkpoint scores $20.3$
points of $F^w_\beta$ below it (Appendix~\S\ref{app:t2full}). EASE and DSS
are reported in a separate block of Table~\ref{tab:t2} because their inference pipelines retain
promptable segmentation models; DSS also uses an MLLM. Our students use a single student forward
pass. These resource differences should be considered when interpreting accuracy and cost, but do
not establish the cause of an accuracy difference. The fixed-pool selection tables evaluate
explicitly labeled scoring adaptations; the complete DSS pipeline is reported separately.

\PAR{The selection rule under a matched training procedure.}
Every downstream row above is supervised by SPHERETRUST's own ranking, so it measures the pipeline
rather than the selector. Appendix~Tab.~\ref{tab:armcmp} therefore trains the three pools under
SPHERETRUST, the size control and the candidate-derived adaptation of the DSS scoring rule. Under
fixed-label training, the three rules separate
by $2.3$ to $5.8$ points, and which rule performs best depends on the regime. The complete
procedure improves all three rules by $2.6$ to $10.7$ percentage points relative to their own
fixed-label baselines and leaves SPHERETRUST first in every regime and every reported seed, with
mean advantages of $0.1$ to $0.4$ points over the size control, which a matched score scale removes
on two of the three regimes (Appendix~\S\ref{app:armcmp}), and $0.2$ to $1.0$ over adapted DSS,
which it does not.
The training procedure therefore improves all three tested scoring rules, with gains varying by
rule and regime. Our downstream results demonstrate the performance of the combined selection and
training pipeline; the total gain does not isolate the contribution of spherical geometry.

\PAR{Generators and anchors.}
SAM3~\citep{carion2025sam3} as both anchor and generator gives a second camouflage pool whose
student gains on every metric of COD10K, NC4K, and CHAMELEON and $4.0$ points of $F^w_\beta$ under
low light, while the anchored pool stays $0.9$ points higher on CAMO
(Appendix~Tabs.~\ref{tab:pools}a and \ref{tab:t2cod}).

\section{Discussion}
\label{sec:disc}

\PAR{Summary.} A pseudo mask read as a partition of the frozen hypersphere carries a label-free
measure of its quality, and the same sphere also supports training. As a selector, SPHERETRUST
exceeds the strongest evaluated external comparator on six pools by $1.7$ to $9.3$ percentage points
in selected Dice and is within $0.1$ percentage points of the candidate-derived DSS adaptation on
the two prompted camouflage pools, with a lower catastrophic-error rate. These comparisons include
explicitly labeled adaptations of related methods. The candidate set, prototype term, and
cross-fitted second round together add $4.5$, $2.3$, and $5.5$ points of weighted $F$ under our
ranking and $2.6$ to $10.7$ points across the three tested rankings.

\label{sec:limstart}\PAR{Limitations.} Selection needs two viable candidates per image, and the
cues operate at the ViT patch scale, so the rule is blind to boundary detail inside a patch, which
dichotomous benchmarks score. SAM3 replaces the multimodal anchor on camouflage only, where its
pools were stronger; on PlantCamo, its student trails the MLLM anchor student by $9.3$ points of
$F^w_\beta$ (Appendix~\S\ref{app:fresh}).
On unprompted SAM output a size control, area in place of coverage, selects as well or slightly
better (Appendix~\S\ref{app:terms}). On camouflage, the candidate-derived DSS adaptation trains the
better fixed-label student, whereas the complete procedure yields a small observed advantage of
$0.18$ percentage points for SPHERETRUST (Appendix~\S\ref{app:armcmp}). On COD10K the students
trail the strongest published configurations, target identity comes from the pool and was examined
on camouflage only, and LL-COD is synthetic.

\label{sec:bodyend}
\section*{Reproducibility Statement}
Appendix~\S\ref{app:impl} lists the model choices, construction of each candidate pool, scoring
constants, and student training schedule. Equations~\ref{eq:F} to \ref{eq:R} define the candidate score
and selection rule, Eq.~\ref{eq:cs} the candidate set loss, and Eq.~\ref{eq:S} the second round
score. The complete code will be released publicly on GitHub, covering pool construction,
candidate scoring and selection, the validation and fold splits, the prototype fit, and both
training rounds, together with the configuration of every reported run, the archived per candidate
scores, and the scripts that produce each table and figure of this paper. The three seeds of the final student share the round two candidate sets
built from seed 0's fold students, so their standard deviations describe variation at the final
stage, conditional on those sets. The configuration of the training stage was chosen on the $1{,}085$
annotated validation images held out of the three training pools. The reported students were then
trained on the complete pools, those images included, and scored once on the test sets.

\section*{AI Use Statement}
We used AI assistants (large language models) for \textbf{writing and language polishing}, LaTeX
compilation and layout (including the method diagram), and \textbf{elementary code assistance} for
routine scripts used in data loading, batch execution, and result tables. AI assistants were also
used to \textbf{cross-check reported numbers and literature comparisons} against the cited sources
and our experimental records, and feedback obtained this way on the comparison tables, the
description of the iterative training and supervision ablation, and the interpretation of the
selection results was incorporated after verification by the authors. The authors produced all
experiments and reported results, reviewed every retained suggestion against the underlying data and
code, and take full responsibility for the paper.

\bibliography{iclr2027_conference}
\bibliographystyle{iclr2027_conference}

\appendix
\addtocontents{toc}{\protect\setcounter{tocdepth}{2}}
\renewcommand{\contentsname}{Appendices}
\clearpage
\tableofcontents
\clearpage
\setlength{\intextsep}{7pt plus 2pt minus 2pt}
\makeatletter\setlength{\@fptop}{0pt}\setlength{\@fpbot}{0pt plus 1fil}\makeatother
{\small
\section{Protocol}

\subsection{Implementation Details}
\label{app:impl}

Unless stated otherwise, all datasets and experiments use the same scoring and student settings.
Dice is computed between the generator's native resolution mask and the native resolution ground
truth, with the ground truth binarized at $0.5$ and no resampling of either, and the $350\times350$
grid is used only for features and for the patch level terms. Every selection number in the paper
follows this protocol.

\begin{enumerate}
\item \textbf{Backbone.} DINOv2 ViT-B/14 (\texttt{timm}), frozen, with $350 \times 350$ inputs and a
$25\times25$ patch grid. Patch features are read from the last of the four probed blocks $3/6/9/12$,
that is from block $12$ (Appendix~\S\ref{app:design}).
\item \textbf{Generator.} SAM ViT-H. P1 uses automatic mask generation with $24$ points along each
side, predicted IoU threshold $0.86$, stability threshold $0.90$, and crop layers disabled. P2 turns
every anchor box into seven prompt boxes, four expansions moving each side out by $0$, $5$, $12$,
and $25\%$ of the box width and height, clipped to the image, and three jitters at amplitudes $6$,
$10$, and $15\%$ displacing the two corners independently, each offset drawn from
$\mathcal{U}[-a,a]$ times the width or height (\texttt{default\_rng(7+\textit{shard})}).
The decoder is called on each prompt box with
\texttt{multimask\_output=True}, returning three masks, so one anchor box yields $21$ raw
masks, and where an image has two or more anchor boxes the most confident mask of each is joined into
a further union candidate. The filters of item 3 and duplicate removal then apply, leaving
$25$ to $28$ candidates an image. The SAM3 pool of \S\ref{sec:e2}
replaces anchor and generator and prompts its own decoder from every instance box
without expansion or jitter.
\item \textbf{Candidate filters.} Relative area in $[0.01,0.70]$, the fraction of frame pixels covered
($|\partial I\cap M|/|\partial I|<0.5$), and at least three patches on both sides of the mask. The P1
unprompted pool retains the top $12$ candidates by predicted IoU, P2 merges and retains all
candidates across prompt boxes (Tab.~\ref{tab:floors}), and these checks precede every prototype
and score.
\item \textbf{Anchor region, prototypes, modes.} $\Omega$ is the bounding box of the union of the
image's candidates on the patch grid, or the union of the expanded prompt boxes when they exist. The
foreground prototype is the renormalized mean direction of the \emph{core}, the patches present in at
least half the candidates and inside $\Omega$. Below two such patches the threshold
drops to $0.3$, and below two again the core is all of $\Omega$, so it stays populated and
$\phi_{\mathrm{fg}}$ is always a unit direction. The background prototype is the border ring's,
two patches wide. Modes are fitted on the patches of $\Omega$ at least one candidate covers,
with Euclidean $k$-means (\texttt{sklearn}) at $k=\min(4,\,\cdot\,)$ over that support,
\texttt{n\_init}$\,$=$\,$3, \texttt{random\_state}$\,$=$\,$0, and the fitted centroids renormalized
to the sphere afterwards. On unit directions the Euclidean and the cosine assignments differ only
through the centroid norms, so this departs from spherical $k$-means only during the fit. Modes closer to the
foreground prototype than to the background one
are kept (the best aligned mode if none is), each staying in the coverage average.
\item \textbf{Fusion.} $z$ scores within each image, summed with equal weights. A term with standard
deviation at most $10^{-6}$ contributes zero, and otherwise its denominator is $\sigma+10^{-6}$. Exact
ties preserve the input order, which is decreasing predicted IoU. The formula, the equal weights, the
normalization, the border ring, $k$, and the candidate filters are one setting shared by every pool,
dataset, and generator in the paper.
\item \textbf{Student.} A DPT-lite decoder \citep{ranftl2021dpt} on the frozen backbone, trained from
scratch, with two refinement stages conditioned on the image at half and full output resolution
($2.3$M parameters). Training uses 25 epochs, batch size $12$, AdamW \citep{loshchilov2019adamw} at
$2\times10^{-4}$ with a cosine schedule and warm up, BCE$+$Dice, random resized crops, photometric
jitter, cross view consistency, and seeds $\{0,1,2\}$. Every student variant, the selection only
reference included, uses this architecture and schedule, and the two supervision modes differ only
in the loss, a fixed label or the candidate set rule of Eq.~(\ref{eq:cs}).
\item \textbf{Sphere term.} The labeled images of a pool are split into two halves by a fixed
permutation. Spherical $k$-means with $25$ iterations on a random subset of $400{,}000$ patch
directions per side of one half gives $16$ foreground prototypes from patches whose \emph{spatial
mask overlap} $o(p)$, the fraction of patch $p$'s $14\times14$ pixels that the selected mask of
Eq.~(\ref{eq:R}) covers, exceeds $0.7$, and $64$ background prototypes from
patches with $o(p)$ below $0.05$ ($o$ is a per patch spatial quantity, unrelated to the
image level $\Csph$ of Eq.~(\ref{eq:C})). The two sides share the prior mass equally, each prototype carries its
cluster fraction, the concentration is $10$, and $G_i$ is the sigmoid of the log
likelihood ratio for every image of the other half. $\Gamma$ enters the round one order
and Eq.~(\ref{eq:S}) as a $z$ score within the image. For an image in one
half, $G_i$ comes from the $80$ prototype directions fitted on the other half, and no mask of
its own half enters that fit. The fold student of a half trains on the images of that half with
candidate sets ordered by $\STscore+z(\Gamma)$. For the image it later predicts, that image's candidates
and its label stay outside its training, but the ordering of its own training candidates was made
with prototypes fitted on the fold it predicts, so the two folds are not independent and we do not
claim they are (\S\ref{sec:m2}). Tab.~\ref{tab:csabl} reports the variant that drops $\Gamma$ from
both rounds, in which the halves are linked through $P_i$ only, which bounds the effect of removing
this route.
\item \textbf{Candidate sets.} Round one keeps for each training image its candidates in decreasing
$\STscore+z(\Gamma)$, greedily dropping any candidate whose IoU with a kept one reaches $0.8$. The
builder stores up to eight per image, and the first $K=5$ enter training with that score as prior
$\pi_{ik}$. The loss picks
$\argmax_k[\pi_{ik}-\ell_{ik}/T]$ with $T=0.1$ once the first five epochs, which use rank one only,
are over, and the choice is detached from the gradient. Round two trains one round one student per
half and stores each student's soft prediction $P_i$ on the other half at $350\times350$. The round
two set holds the completed mask $U_i$ first, with the prior of the leading candidate under $S$,
followed by the four leading distinct candidates under $S$, and training repeats the rule above, so
its warm up epochs train on $U_i$. The fold students, $P_i$, $G_i$, and the round two sets come from
seed $0$ and are shared by the three seeds of the final student.
\item \textbf{Validation splits and configuration.} Each training pool holds out a fixed tenth of
its images, drawn once with \texttt{RandomState(2027)} over the sorted stems ($386$, $402$, and $297$
for COD, SOD, and DIS, $1{,}085$ in all), kept out of every run that took part in choosing
the configuration, candidate sets, folds, prototype fit, out-of-fold maps and students alike. They
were \emph{not} held out of the reported students, which were trained on the
complete pools once the choice was frozen, which is why Tab.~\ref{tab:floors}'s pool sizes
are the full $3{,}857$, $4{,}025$, and $2{,}969$. $K$, $T$, the warm up length, the deduplication threshold, and the prototype
counts are code constants. Two settings were open, the round one training order and the treatment
of the completed mask in round two (Tab.~\ref{tab:config}). The choice is the highest mean
$F^w_\beta$ over the three validation splits, one configuration for all regimes, with differences
under $.001$ counted as ties and resolved toward the more conservative setting, for the threshold the
one that admits fewer completion pixels, round one first and round two on its fold students. The
chosen configuration was then trained on the full pools with three seeds, with the validation splits
as the only images scored during training, and the test sets were scored once afterwards, which
gives every student number of the paper.
\item \textbf{Anchor boxes.} The anchor provides box prompts when the training pool is built and at
that point only. The MLLM anchor is Qwen2.5-VL-7B-Instruct \citep{bai2025qwen25vl} with
greedy decoding and at most three boxes. Two prompts per image ask for the single most prominent
foreground object and for every distinct salient object, with wording for each regime
(Appendix~\S\ref{app:mllm}), the box sets are merged, and duplicates with IoU above $0.9$ are
removed. The SAM3 anchor \citep{carion2025sam3} grounds each phrase of a short ladder separately
(Appendix~\S\ref{app:mllm}) and prompts its own mask decoder from every instance box. \end{enumerate}

\subsection{Candidate Pools, Eligibility, and Datasets}
\label{app:floors}
\label{app:regime2}
P1 is the unprompted SAM construction capped at the top $12$ candidates by predicted IoU. P2 merges
candidates generated from the anchor box prompts and retains the complete merged set. Two prompted
constructions appear. The single prompt one, from the first anchor prompt only ($3{,}743$,
$4{,}019$, and $2{,}947$ images, about $20$ candidates each), serving the
diagnostics of Appendix~\S\ref{app:diag}, and the merged
one from both prompts ($3{,}857$, $4{,}025$, and $2{,}969$ images, $25$ to $28$ candidates
each), which Tab.~\ref{tab:floors} reports and every student of Tabs.~\ref{tab:csabl},
\ref{tab:t2}, \ref{tab:pools}, and \ref{tab:csabl} trained on.
The three source splits are of unequal size, so each is capped before any candidate is generated. The
camouflage split (CAMO-train $+$ COD10K-train) holds $4{,}040$ images and DIS-TR $3{,}000$, both taken
whole, while DUTS-TR's $10{,}553$ are cut to the first $4{,}040$ stems in sorted filename order. The
cut is a deterministic prefix rather than a random subsample, so it needs no seed and is reproducible from
the dataset alone. The eligibility filter then leaves
$3{,}857$, $4{,}025$, and $2{,}969$ images. The index lists are released with checksums. Each pool is built once before
any selector is applied. The two families differ in baseline candidate
quality (Tab.~\ref{tab:floors}). P1 candidates vary more widely in target identity and coverage,
which is why P1 serves selection analysis and P2 training, and the two contain different images.
On the single prompt construction of P2, with roughly $20$ largely overlapping candidates per image
against about $\pfSelKCod$ under P1, generator confidence and SPHERETRUST both select a higher
Dice than on P1, at a lower catastrophic rate and a lower Top-1, and SPHERETRUST leads on both
families (Tabs.~\ref{tab:t1} and \ref{tab:t1unp}).

\PAR{The SAM3 pool.}\label{app:sam3sel}
Tab.~\ref{tab:t1} runs every selector on the pool SAM3 builds alone (\S\ref{sec:e2}), $3{,}787$
COD training split images ($2{,}833$ from COD10K-train, $954$ from CAMO-train) with $56{,}705$
candidates, under the eligibility filter below. Generator confidence is SAM3's own score for the
mask and generator stability the IoU of its two thresholded logit maps, and a candidate formed as the
union of two boxes ranks last under both.

\begin{table}[!htb]
\caption{\textbf{Selection pools and training pools.} $k$ is candidates per image, and \emph{random} and
\emph{oracle} define the available selection gap. Selection uses unprompted generator output, training
pools prompt the same generator from the MLLM anchor (the SAM3 pool is in Tabs.~\ref{tab:t1} and \ref{tab:pools}), and
LL-COD reuses the degraded COD10K test set~\citep{fan2020camouflaged} with the camouflage students.}
\label{tab:floors}
\apptab
\begin{tabular}{lrrrr rrrr}
\toprule
& \multicolumn{4}{c}{selection pool (Tab.~\ref{tab:t1unp})} & \multicolumn{4}{c}{training pool (Tab.~\ref{tab:t2})} \\
\cmidrule(lr){2-5}\cmidrule(lr){6-9}
regime & $n$ & $k$ & random & oracle & $n$ & $k$ & random & oracle \\
\midrule
Camouflage  & \pfSelNCod   & \pfSelKCod   & \pfSelRandomCod   & \pfSelOracleCod   & \pfTrnNCod   & \pfTrnKCod   & \pfTrnRandomCod   & \pfTrnOracleCod \\
Salient     & \pfSelNSod   & \pfSelKSod   & \pfSelRandomSod   & \pfSelOracleSod   & \pfTrnNSod   & \pfTrnKSod   & \pfTrnRandomSod   & \pfTrnOracleSod \\
Dichotomous & \pfSelNDis   & \pfSelKDis   & \pfSelRandomDis   & \pfSelOracleDis   & \pfTrnNDis   & \pfTrnKDis   & \pfTrnRandomDis   & \pfTrnOracleDis \\
Low light   & \pfSelNLl    & \pfSelKLl    & \pfSelRandomLl    & \pfSelOracleLl    & \multicolumn{4}{c}{\emph{direct camouflage evaluation}} \\
\bottomrule
\end{tabular}
\end{table}

\begin{figure}[!htb]
\centering
\includegraphics[width=0.8\linewidth]{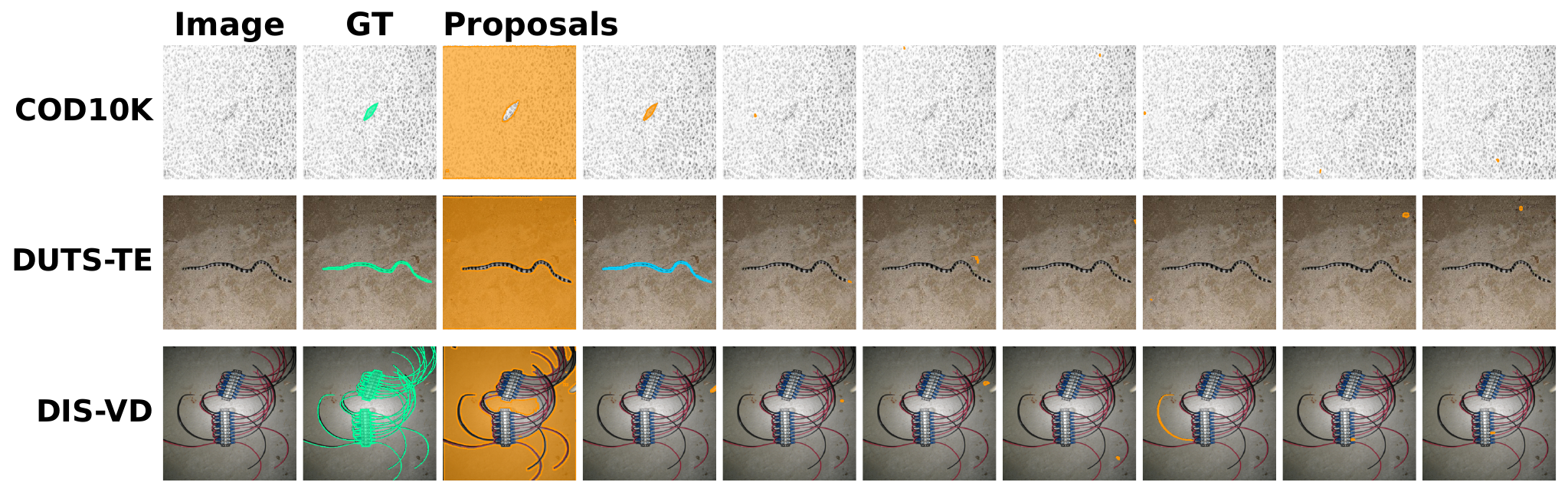}
\caption{\textbf{Images with fewer than two eligible candidates} after the plausibility filters, one
per regime, with the ground truth and every proposal the generator returned.}
\label{fig:starve}
\end{figure}

\PAR{Pool eligibility.} Selection is evaluated on images whose annotation contains both foreground
and background and that have at least two candidates passing the common filters, with candidates
generated before ground truth is consulted. Eligible counts are COD10K $1979/2026$
($97.7\%$), DUTS-TE $4993/5019$ ($99.5\%$), DIS-VD $465/470$ ($98.9\%$), and LL-COD $1932/2026$
($95.4\%$), and downstream uses the full test sets. The filters of
Appendix~\S\ref{app:impl} apply to candidates and identically for every selector, so an image
whose target is smaller than the $1\%$ area floor stays eligible whenever the generator
returns two admissible candidates elsewhere (Fig.~\ref{fig:starve}).

\PAR{Datasets and regimes.}
The $15$ evaluation sets are $14$ public benchmarks plus LL-COD, four camouflage sets
\citep{le2019anabranch,fan2020camouflaged,lv2021simultaneously,skurowski2018animal} (CAMO, COD10K,
NC4K, CHAMELEON), five salient sets
(ECSSD, DUTS-TE, DUT-OMRON, HKU-IS, PASCAL-S)
\citep{yan2013ecssd,wang2017duts,yang2013dutomron,li2015hkuis,li2014pascals}, five DIS splits (DIS-VD
and DIS-TE1 to DIS-TE4) \citep{qin2022dis}, and LL-COD, a synthetic condition on COD10K test images.
With seed $13$, P1 samples $1157$ images from the standard $4040$-image COD (CAMO-train $+$
COD10K-train) split, disjoint from testing, and the merged P2 contains \pfTrnNCod, \pfTrnNSod, and
\pfTrnNDis\ images for COD, SOD, and DIS after filtering, with checksummed lists. The target
ambiguity diagnostic of \S\ref{sec:e1} uses the unprompted SAM pool of the whole $4{,}040$-image
split, of which $3{,}925$ images pass the eligibility filter above, with $10.1$ candidates per
image. LL-COD
applies to every COD10K test image a gamma of $2.5$, an exposure factor of $0.45$, shot noise of
$0.7\sqrt{\text{signal}}$, read noise with $\sigma=6$, and Gaussian blur with $\sigma=0.6$ at seed
$0$, leaves the ground truth unchanged, and every compared method is rerun on the degraded images
(Selfment as in Appendix~\S\ref{app:t2full}, A2S-v3 with its released stage two checkpoint at $320$
pixels, CSNet at $352$ pixels followed by dense CRF).

\subsection{Prompts and Phrase Ladders}
\label{app:mllm}

\PAR{Anchor prompt (pool construction).}
{\footnotesize
\emph{``You are given an image containing a camouflaged object. The object may have similar color,
texture, or illumination to the background. Identify the most likely camouflaged foreground object.
Return ONLY up to 3 bounding boxes in JSON as a list:
\texttt{[\{"bbox\_2d":[x1,y1,x2,y2],"confidence":0.0\}]}. Coordinates in pixels of THIS image. Boxes
should tightly cover the full visible object. Do not select ordinary background.''}
This is the exact first anchor prompt for camouflage. The second prompt, used for every regime, is

\emph{``You are given an image. Identify EVERY distinct salient foreground object in it, not only the
most prominent one. Objects may be subtle or low-contrast against their surroundings. If there is
genuinely only one salient object, return exactly one box. Return ONLY bounding boxes in JSON as a
list: \texttt{[\{"bbox\_2d":[x1,y1,x2,y2],"confidence":0.0\}]}. Coordinates in pixels of THIS image.
One box per object, tightly covering that object's full visible extent. Do not merge several objects
into one box, and do not select ordinary background.''}

and the first prompt outside camouflage, which the salient and dichotomous pools use, is

\emph{``You are given an image containing one main foreground object or region of interest, which may
be subtle or low-contrast against its surroundings. Identify the single most prominent foreground
object. Return ONLY up to 3 bounding boxes in JSON as a list:
\texttt{[\{"bbox\_2d":[x1,y1,x2,y2],"confidence":0.0\}]}. Coordinates in pixels of THIS image. Boxes
should tightly cover the full visible object. Do not select ordinary background.''}
}

Decoding is greedy in every case, and the box expansion and jitter that turn these boxes into prompt
boxes are in Appendix~\S\ref{app:impl}, item 2.

\PAR{SAM3 phrase ladders.}
The SAM3 anchor grounds four phrases per regime, one at a time, and keeps every instance box together
with the index of its phrase. The ladder is \emph{animal, camouflaged animal, insect, object} for camouflage,
\emph{object, person, animal, vehicle} for salient objects, and \emph{object, animal, vehicle, tool}
for dichotomous segmentation. The camouflage ladder gives the pool of
Tab.~\ref{tab:pools}, and the salient and dichotomous ladders gave weaker students than the MLLM
anchor, so those regimes keep it.

\section{What the Selection Measures Capture}

\subsection{What Each Term Contributes}
\label{app:terms}

\PAR{Term ablations.} These use the P1 camouflage training subset
(\ablMeta) unless stated otherwise, whereas Tab.~\ref{tab:t3a} uses the three prompted training
pools, on which the margin of the full score over the $\Fcon+\Ebd$ pair is $+.008$,
$+.037$, and $+.097$, and $+.033$, $+.111$, $+.121$, and $+.030$ on the four unprompted test pools.
On the training subset, removing angular contrast, the
boundary prior, and spherical coverage changes selected Dice by \ablDropF, \ablDropE, and \ablDropC,
the boundary prior alone selects at \ablE\ against \ablFull\ (\ablDE, CI
\ablCiE), and subtracting $z(\Ebd)$ from generator confidence, TokenCut, and perturbation
consistency yields \ablSamE, \ablTokE, and \ablPcE. The $\Fcon+\Ebd$ pair, which shares features,
candidates and normalization with the full score, sits
\ablDHeu\ below the full score here (CI \ablCiHeu), with margins of \ablPoolsRange\ across four pool
constructions of the released ablation data, each a macro average over
seven object datasets, positive on \ablPoolsN, and of changing sign on four medical sets
outside this paper's scope.

\begin{figure}[!htb]
\centering
\includegraphics[width=0.66\linewidth]{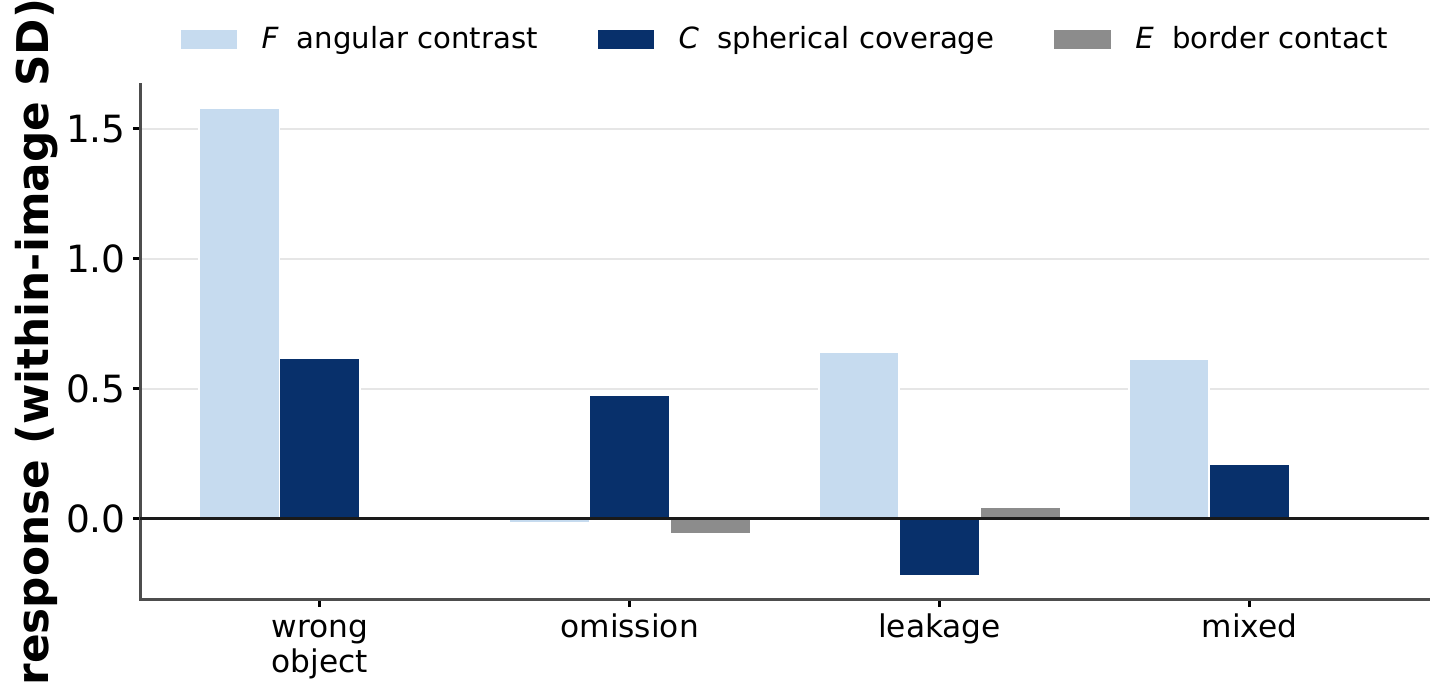}
\caption{\textbf{How the cues respond to errors.} Standardized responses of the three cues to
controlled corruptions of the ground truth mask, grouped into wrong object, omission, leakage, and
mixed errors.}
\label{fig:curves}
\end{figure}

\PAR{Responses to controlled corruptions.}
Fig.~\ref{fig:curves} groups seven controlled corruptions of the ground truth mask (translation by
$5$ to $40\%$ of the diagonal, substitution of another image's object, erosion and dilation with
kernels of $1$ to $13$ pixels, cropping to $80$ down to $20\%$, a background blob of $5$ to $40\%$
relative area, and a mix of translation, erosion, and blob) into errors of object identity, omission,
leakage, and mixed families, each at seven levels. Each response is the score of the clean mask minus
that of the corrupted mask, normalized by the term's standard deviation within the image. $\Fcon$
changes most for object identity and leakage, $\Csph$ responds to omissions and negatively to
leakage, so coverage alone can favor oversized masks, and $\Ebd$ changes little because these
corruptions rarely reach the image boundary.

\PAR{Area as a surrogate that depends on the pool.} Let $a_{\mathrm{mask}}(M)=|M|/(H_iW_i)$ denote the
fraction of image pixels covered by mask $M$. Table~\ref{tab:sizecontrol} compares the full score
$z(\Fcon)+z(\Csph)-z(\Ebd)$ with the area replacement $z(\Fcon)+z(\log a_{\mathrm{mask}})-z(\Ebd)$ on
identical candidates, over the P1 camouflage macro (CAMO, COD10K, NC4K, and CHAMELEON), the prompted
COD, SOD, and DIS training pools, and the DPT-lite pool of Tab.~\ref{tab:pools}, with $20{,}000$
paired bootstrap draws over images within each dataset and equal weight to the datasets of a family.
The sign changes with the pool, in favor of area on unprompted SAM output and of coverage on prompted
SAM output and on the pool from another generator. The mean also hides that the two fail differently. Area has the higher Top-1 on every prompted pool, by $2.8$, $3.3$, $14.9$, and $2.6$ points, and the
higher catastrophic rate on all four, $.089$ against $.073$ on the camouflage SAM pool and $.024$
against $.010$ on the salient one, because rewarding size picks the best candidate more often when
proposals are fragmented and leaks badly when they are intact, whereas coverage stops rewarding a mask
once it spans the foreground modes. Coverage also assumes a localized anchor, since its foreground
prototype is the mean direction of the core, the patches at least half the candidates cover,
and its modes are fitted on the candidate union inside $\Omega$, so under a prompted pool, where
candidates concentrate on one or a few boxes, the core is a reliable proxy for the target's interior,
while under unprompted output, spread over the whole frame, it is small and often
falls on background clutter and the modes then belong to that clutter. Area needs no such anchor, which is
why it survives the change of pool that undoes coverage (Tab.~\ref{tab:t1unp}). Coverage nonetheless stays associated with true Dice
after conditioning on the other terms and area, by regressing, within each image, the ranks of $\Csph$
and of true Dice on the ranks of $\Fcon$, $\Ebd$, and $\log a_{\mathrm{mask}}$ (images with at
least eight candidates) and correlating the residuals gives partial Spearman coefficients of
$+.382\;[+.348,+.413]$ on the unprompted macro, $+.242\;[+.234,+.251]$ on the prompted macro, and
$+.112\;[+.080,+.145]$ on the DPT-lite pool, and among candidates of one image whose areas differ by
at most five percent, $\Csph$ picks the mask with greater Dice in $\xaPairCCod$, $\xaPairCDuts$, and
$\xaPairCDis$ of pairs on COD10K, DUTS-TE, and DIS-VD, and area in $\xaPairACod$, $\xaPairADuts$,
and $\xaPairADis$ of the same pairs.

\begin{table}[!htb]
\caption{\textbf{Coverage and area on identical candidates.} $\Delta$ is selected mask Dice for
$z(\Fcon)+z(\Csph)-z(\Ebd)$ minus that for $z(\Fcon)+z(\log a_{\mathrm{mask}})-z(\Ebd)$, with a
$95\%$ bootstrap interval.}
\label{tab:sizecontrol}
\apptab
\begin{tabular}{lrrrl}
\toprule
pool family & datasets & images & candidates & $\Delta$ selected Dice [95\% CI] \\
\midrule
unprompted SAM macro & 4 & $6{,}340$ & $64{,}962$ & $-.0150\;[-.0192,-.0107]$ \\
prompted SAM macro & 3 & $10{,}851$ & $283{,}759$ & $+.0206\;[+.0178,+.0235]$ \\
non-SAM pool & 4 & $913$ & $10{,}338$ & $+.0045\;[+.0004,+.0089]$ \\
\bottomrule
\end{tabular}
\end{table}

\subsection{Analysis of the Scoring Rule}
\label{app:design}

\PAR{Directional features.}
For a single von Mises--Fisher component per class \citep{mardia2000directional,banerjee2005clustering}
with common concentration, the foreground-to-background log-likelihood ratio is linear in the
normalized feature direction. With multiple components, it becomes a difference of log-sum-exp terms
and is generally nonlinear. This motivates using feature direction, but does not establish angular
contrast as an estimator of mask quality. We compare particular direction-based and magnitude-based
statistics empirically in \S\ref{sec:e1} and below; these comparisons do not determine whether
magnitude adds information beyond direction. Equation~\ref{eq:F} normalizes each patch before
averaging and therefore discards patch magnitudes. Averaging raw features generally changes the
mean directions and the resulting contrast. Both implementations nevertheless achieve the same
reported AUROC of $.946$ on the unprompted COD10K pool in this experiment. We also compare a
median-margin statistic in normalized and raw-feature forms. With $\bar f_{\mathrm{in}}$ and $\bar f_{\mathrm{out}}$ the mean patch
feature inside and outside a mask, the margin of patch $p$ is
$q_p=\langle f_p,\bar f_{\mathrm{in}}\rangle-\langle f_p,\bar f_{\mathrm{out}}\rangle$, and the
median margin is the median of $q_p$ over inside patches minus its median over outside patches. On
unit directions, with $\fhat_p$ in place of $f_p$ and both means normalized, it reaches AUROC
$.934$, on the raw features $.899$, and the feature norm alone, the mean $\lVert f_p\rVert_2$ inside
minus outside, $.752$. Partialling out area reduces the Spearman coefficient of angular contrast by
$0.015$, and under photometric perturbations of $59{,}720$ candidate pairs the angular score flips
$3.0\%$ of rankings against $5.0\%$ for the variant with magnitude.

\PAR{Layer and backbone.}
Across blocks $3/6/9/12$ of the frozen backbone on the P1 camouflage training subset (\ablMeta),
selected Dice from angular contrast rises from $.162$ to $.575$ ($.171$ and $.475$ at blocks $6$ and
$9$) and AUROC from $.525$ to $.938$ ($.570$ and $.900$ at the two middle blocks), and the same
monotone pattern holds on a COD10K test pool ($\lyrNcand$ candidates over $\lyrNimg$ images,
$\lyrDiceThree$ to $\lyrDiceOneTwo$). The subset's oracle is $.685$.
Block $12$, the final block of the backbone, is the deployed setting. On the P1 pool ($\bbNcand$
candidates), the AUROC of angular contrast is $\bbBauroc$ with DINOv2 ViT-B, $\bbLauroc$ with ViT-L,
and $\bbSauroc$ with SAM's image encoder, alongside the selected Dice reported in \S\ref{sec:e1}.

\PAR{Normalization and candidate count.} On the pool of Appendix~\S\ref{app:learned}, scaling each
statistic to $[0,1]$ within the image in place of $z$ scores changes selected Dice by $0.001$, whereas ranks cost $0.030$,
since ranks discard the spacing between candidate scores within an image. Selection quality increases
with the number of candidates ($3/5/8/12 \to 0.494/0.572/0.623/0.650$).

\section{Detailed Results for Candidate Selection}

\subsection{Stress Test on Unprompted SAM Output}
\label{app:unprompted}

Tab.~\ref{tab:t1unp} repeats Tab.~\ref{tab:t1} on the four unprompted P1 test
pools, one per regime, where SAM proposes every segment and the twelve candidates with
the highest predicted IoU are kept. The average candidate has Dice $\pfSelRandMin$ to
$\pfSelRandMax$ with the target, the oracle reaches $\pfSelOracleDis$ to
$\pfSelOracleSod$, and $91\%$ of camouflage training images contain a decoy, so
a selector must first find the target among unrelated segments and then judge its extent.
SPHERETRUST leads the strongest evaluated external comparator, which on all four of these pools is
the candidate-derived DSS adaptation, by $+9.3$ points of selected Dice on DUTS-TE, $+4.9$ on DIS-VD, $+2.5$ on COD10K, and
$+3.1$ on LL-COD, with the highest Top-1 and the lowest catastrophic rate on every pool.
SelfMask's consensus recovers $.087$ to $.288$ of the gap between random selection and the
oracle here, against $.644$ to $.868$ for $\STscore$. The size control selects better than the full
score on three of the four pools ($+1.2$, $+1.7$, and $+1.3$ points, $-0.3$ on DIS-VD), the reverse
of the prompted, SAM3, and decoder pools, and Appendix~\S\ref{app:terms}
examines why area is the stronger single surrogate on unprompted output.

\begin{table}[!htb]
\caption{\textbf{Unsupervised selection on unprompted SAM output}, one fixed pool per regime. Marks
are those of Tab.~\ref{tab:t1}, and the rows below the rule are references and carry none.
Appendix~\S\ref{app:floors} states eligibility.}
\label{tab:t1unp}
\apptab
\renewcommand{\arraystretch}{0.88}
\begin{tabularx}{\linewidth}{@{}lYYYY@{}}
\toprule
Selector & \textbf{COD} & \textbf{SOD} & \textbf{DIS} & \textbf{LL} \\
\multicolumn{1}{@{}l}{} & \footnotesize\itshape COD10K & \footnotesize\itshape DUTS-TE & \footnotesize\itshape DIS-VD & \footnotesize\itshape LL-COD \\
\midrule
\multicolumn{5}{@{}l@{}}{\emph{each entry:}~~sel-Dice$\uparrow$ / Top-1$\uparrow$ / Dice$<$.2$\downarrow$} \\
\addlinespace[1pt]
Random & .132/.162/.821 & .170/.129/.743 & .099/.120/.824 & .131/.183/.813 \\
SelfMask vote~\citep{shin2022selfmask} & .218/.221/.723 & .348/.274/.534 & .133/.110/.757 & .203/.219/.731 \\
SAQ~\citep{lin2024saq} & .102/.133/.855 & .141/.098/.777 & .072/.067/.867 & .082/.131/.876 \\
SAM predicted IoU~\citep{kirillov2023segment} & .256/.303/.693 & .285/.264/.650 & .096/.131/.843 & .201/.271/.741 \\
SAM stability~\citep{kirillov2023segment} & .147/.175/.806 & .181/.143/.745 & .078/.071/.862 & .128/.179/.819 \\
TokenCut~\citep{wang2022tokencut} & \third{.518}/\third{.656}/\third{.384} & \third{.543}/\second{.619}/.364 & \third{.283}/\second{.445}/\third{.563} & \third{.444}/\third{.622}/\third{.435} \\
\begin{tabular}[c]{@{}l@{}}Generator confidence with\\UCOD-MKD-inspired structural filtering\end{tabular} & .372/.399/.558 & .508/.444/\third{.361} & .190/.176/.660 & .317/.367/.595 \\
Adapted DSS scoring rule (candidate-derived map) & \second{.580}/\second{.677}/\second{.317} & \second{.574}/\third{.562}/\second{.325} & \second{.299}/\third{.387}/\second{.536} & \second{.490}/\second{.632}/\second{.390} \\
\rowcolor{ourfill}
\textcolor{rkbest}{\textbf{SPHERETRUST}} & \best{.605}/\best{.723}/\best{.278} & \best{.667}/\best{.700}/\best{.202} & \best{.348}/\best{.493}/\best{.456} & \best{.521}/\best{.681}/\best{.336} \\
\midrule
\emph{size control:} $z(\Fcon)+z(\log\text{area})-z(\Ebd)$ & .618/.742/.268 & .684/.733/.185 & .345/.488/.458 & .533/.711/.324 \\
oracle & .677/1.000/.195 & .788/1.000/.066 & .486/1.000/.222 & .603/1.000/.235 \\
\bottomrule
\end{tabularx}

\end{table}

\subsection{Detailed Selector and Student Comparisons}
\label{app:diag}
\label{app:students}

Tab.~\ref{tab:diag} evaluates every selector on the P1 camouflage pool under one protocol, and
Tab.~\ref{tab:students} trains a student with each spherical term alone and with the full score,
the supervision held fixed at one label per image as in the selection only row of
Tab.~\ref{tab:csabl}, on a $3{,}744$ image construction of the camouflage training pool, so its $.772$ mDice and the $.7513$ $F^w_\beta$ of Tab.~\ref{tab:csabl} describe
different pools and metrics. Selected Dice is likewise pool specific, $0.605$ in
Tab.~\ref{tab:t1unp} on \pfSelNCod\ COD10K test images, $0.629$ in
Tab.~\ref{tab:diag} on the $1{,}157$ image training subset, $0.650$ in
and Appendix~\S\ref{app:learned} on the macro pool over four sets. Timing in Tab.~\ref{tab:diag}
starts after candidate generation and the shared DINOv2 forward pass.

\PAR{Consensus, contour, and perturbation baselines.}
The consensus rows of Tab.~\ref{tab:t1} and Tab.~\ref{tab:t1unp} use SelfMask's rule
\citep{shin2022selfmask} in full, dropping the candidates
whose bounding box spans the frame ($\xmFrameMin$ to $\xmFrameMax$ of them) and taking the
highest mean pairwise IoU among the rest. It is reported on those eight pools rather than on the
fixed pool of Tab.~\ref{tab:diag}.
SAQ \citep{lin2024saq} scores the predicted contour
alone with the released implementation, once per image on the shared crop.
\emph{Perturbation consistency} is a control of ours rather than a published selector. The frozen DINOv2
is run on four perturbations of the image (horizontal flip, gamma $0.6$ and $1.6$, Gaussian
blur), the normalized cut foreground of each is realigned to the original frame, and their
average gives a saliency map $S_{\mathrm{pc}}\in[0,1]$ that is stable under perturbation. A candidate scores
$\mathrm{mean}(S_{\mathrm{pc}}\mid M)-\mathrm{mean}(S_{\mathrm{pc}}\mid \bar M)$, at five
backbone passes and five cuts an image, $31$\,s against $0.55$\,s for $\STscore$.

The structural checks alone lift the generator's own confidence from $.276$ to $.400$ selected Dice
on this pool, while the agreement statistic they are paired with ranks candidates no better than
chance (AUROC $.459$). On the prompted pools of Tab.~\ref{tab:t1}, where the candidates of an image
are near duplicates, the same row reaches $.715$ to $.889$.

Fig.~\ref{fig:qualapp} shows the images on which $\STscore$ and generator confidence disagree most.

\begin{figure}[!htb]
\begin{minipage}[c]{0.64\linewidth}
\apptab
\setlength{\tabcolsep}{2.5pt}
\centering
\begin{tabular}{l cccccr}
\toprule
selector & AUROC & sel.\ Dice & regret$\downarrow$ & AURC$\downarrow$ & Top-1 & cost \\
\midrule
\multicolumn{7}{l}{\emph{full pool: 1157 images, 11848 candidates}} \\
SAM predicted IoU & .623 & .276 & .410 & .723 & .318 & $0$ \\
$\Fcon+\Ebd$ pair $z(\Fcon)-z(\Ebd)$ & .936 & .591 & .094 & .280 & .596 & $<1$\,ms \\
TokenCut & .919 & .544 & .142 & .349 & .682 & $5.9$\,s \\
perturbation consistency & .929 & .586 & .099 & \best{.207} & .714 & $31$\,s \\
SAQ & .470 & .098 & .587 & .955 & .126 & $0.31$\,s \\
\begin{tabular}[c]{@{}l@{}}Generator confidence with\\UCOD-MKD-inspired filtering\end{tabular} & .663 & .400 & .285 & .515 & .424 & $0.5$\,s \\
\begin{tabular}[c]{@{}l@{}}Adapted DSS\\(candidate-derived map)\end{tabular} & \best{.956} & .590 & .096 & .212 & .684 & $<1$\,ms \\
\midrule
\textbf{SPHERETRUST} & .955 & \best{.629} & \best{.056} & .232 & \best{.754} & $0.55$\,s \\
\midrule
oracle & -- & .685 & .000 & .092 & 1.000 & -- \\%
\bottomrule
\end{tabular}
\captionof{table}{\textbf{Every selector on one fixed pool.} P1 retains up to twelve masks per image
from the COD training split, ordered by generator confidence. AUROC treats Dice $\ge0.5$ as positive,
cost is per image on one RTX A5000 with shared features, and the AURC mark is on the lowest
value. Single terms give AUROC $.938$ ($\Fcon$) and $.856$ ($\Csph$). The candidate-derived
adaptation of DSS is included as a separate row. \best{Red} marks the best observed value among the
reported selection rules for each metric. The consensus baseline is reported on the eight pools of
Tabs.~\ref{tab:t1} and \ref{tab:t1unp} rather than here.}
\label{tab:diag}

\vspace{1.1\baselineskip}
\apptab
\centering
\begin{tabular}{lcc}
\toprule
selection rule & student mDice & seeds \\
\midrule
train on the candidate union & .553$\pm$0.001 & 3 \\
SAM predicted IoU & .748$\pm$0.003 & 3 \\
$\Csph$ only & .684$\pm$0.002 & 3 \\
$\Fcon$ only & .753$\pm$0.001 & 3 \\
SPHERETRUST, $z(\Fcon){+}z(\Csph){-}z(\Ebd)$ & \best{.772}$\pm$0.002 & 3 \\
oracle (GT ceiling) & .802$\pm$0.001 & 3 \\%
\bottomrule
\end{tabular}
\captionof{table}{\textbf{Selection controls and individual terms in downstream training} (mean
mDice over four COD sets and three seeds, with $\pm$ the seed standard deviation). Only the mask
selection rule changes between rows.}
\label{tab:students}
\end{minipage}\hfill
\begin{minipage}[c]{0.34\linewidth}
\centering
\includegraphics[width=\linewidth]{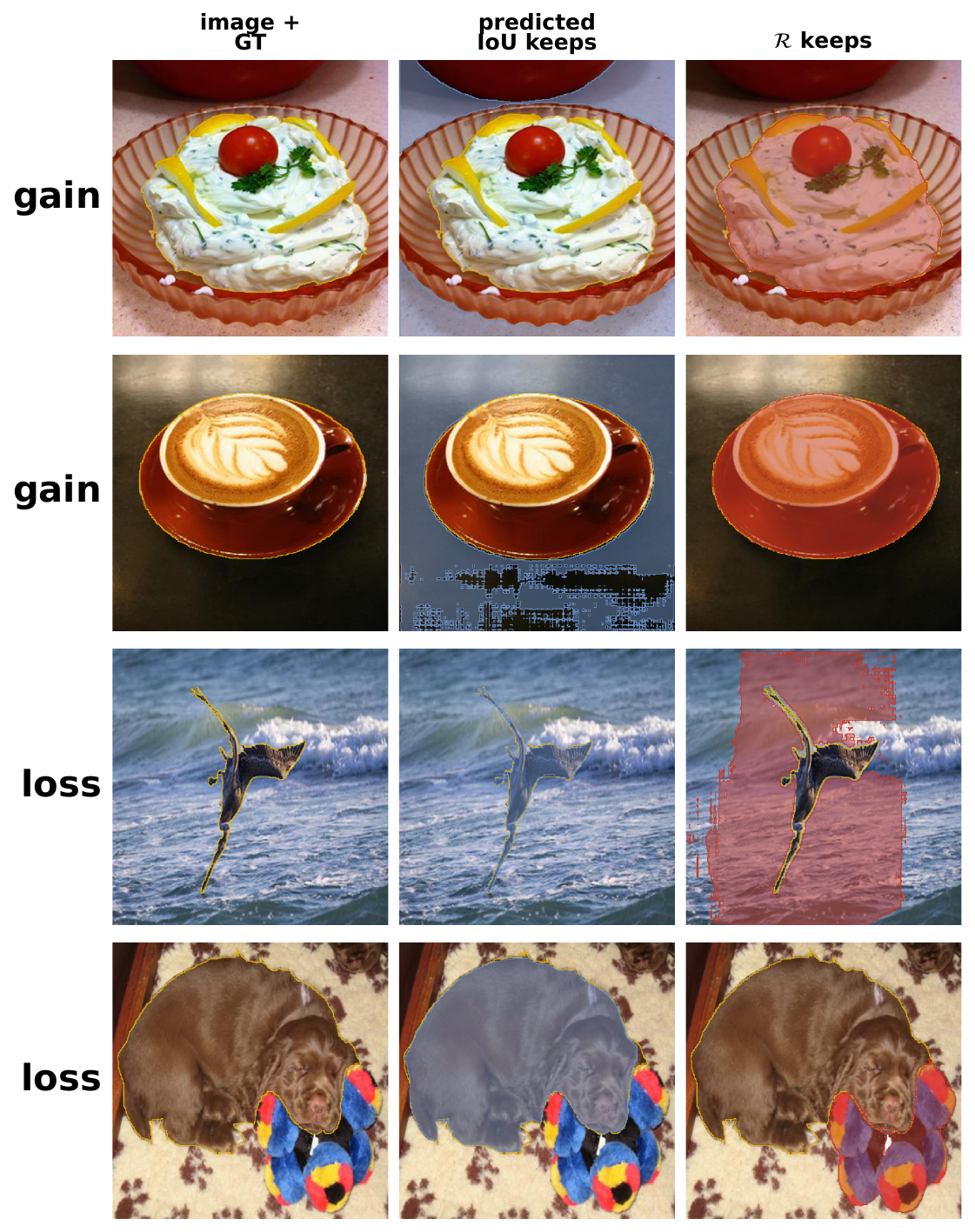}
\caption{\textbf{Largest disagreements between $\STscore$ and predicted IoU.} These are examples, while
the decoy statistics of \S\ref{sec:e1} are computed over the whole unprompted pool of the COD
training split. Columns show ground
truth and each selector's choice. Gains recover the target from a confident but disjoint region.
Losses illustrate the complement symmetry $\Fcon(1-M)=\Fcon(M)$, where the partition and the
annotated object disagree.}
\label{fig:qualapp}
\end{minipage}
\end{figure}

\subsection{The Two Closest Selectors on a Fixed Pool}
\label{app:closest}

We compare scoring adaptations inspired by two related methods on identical candidate pools. These
adaptations are distinguished from the original methods and their complete pipelines.

\PAR{Adapted DSS scoring rule (candidate-derived map).} DSS \citep{yang2026dss} discovers coarse regions by clustering frozen
features, reads a similarity map from each region, prompts SAM2 with a box from that map, and scores
the resulting mask by Eq.~(7), $s_i=\mathrm{corr}(m_i,\mathrm{sim}_i)+\big(1-\mathrm{BC}(m_i)\big)$,
with $\mathrm{corr}$ the Pearson correlation between the mask and the similarity map of the region
that prompted it and $\mathrm{BC}$ the fraction of a border strip the mask occupies. Its selection is
therefore defined on the candidates its own discovery stage produces, and on a pool built elsewhere
the map it reads is absent. The row applies the formula to our candidates with two
substitutions, stated here so the row is read for what it is. The similarity map is the cosine
between every patch direction and the mean direction of the patches the candidate itself covers, on
the frozen DINOv2 ViT-B/14 every other row uses, and $\mathrm{BC}$ is the border term $\Ebd$ of
Eq.~(\ref{eq:E}). The row therefore compares scoring rules on one backbone and one pool, and DSS as a
pipeline, with its own discovery, SAM2 and multimodal stages, is compared as a whole in
Tab.~\ref{tab:t2} and Tab.~\ref{tab:t2cod}.

We additionally evaluate an anchor-box adaptation, deriving the similarity map from the patches of
the anchor box that generated the candidate, rasterized on the same $25\times25$ grid and read
through the same foreground routine. The generating box is recorded for candidates from images with
a single box, which is $47\%$ to $69\%$ of each pool, and assigned otherwise by the box holding the
largest share of the mask. The reference map therefore uses the box metadata from pool construction
without using the candidate mask being scored, which is what DSS does for its own prompts. Both
variants substitute the same border term and both are reported in Tab.~\ref{tab:t1}. The anchor-box
map selects $.620$, $.549$, $.839$, and $.606$ against $.825$, $.788$, $.912$, and $.720$ for the
candidate-derived map, below random selection on both camouflage pools, because the mean direction
of a box mixes the object with the background the box also contains, and two other assignment
rules, the box whose geometry best matches the mask and the union of the image's boxes, move the row
by at most $2.9$ points against the $7$ to $24$ points that separate it from the candidate-derived
map. The margins reported in \S\ref{sec:e1} therefore use the stronger of the two adaptations, and
are a conservative comparison across two adaptations of the scoring rule rather than the performance
of one fixed selector.

\PAR{Generator confidence with UCOD-MKD-inspired structural filtering.} In the released UCOD-MKD
procedure \citep{chen2026beyond}, the candidate with the highest SAM predicted IoU is selected
within each prompt group before grading. Agreement statistics and structural checks then determine
a quality tier for distillation. The structural checks grade the selected mask in the intermediate
agreement case; they do not filter the candidate pool before selection. Our comparison instead
applies a structural filter before selecting the highest-confidence remaining candidate. We retain
the ten-pixel edge margin, twenty-pixel strip threshold, at most one truncated edge, and at most
$49$ connected components. This adaptation rejects $15\%$ to $31\%$ of candidates. On the merged
camouflage pool it improves selected Dice from $.728$ to $.772$; on the SAM3-only pool the gain is
$0.2$ points. These results describe the adaptation, not the original selection procedure.

\PAR{What the comparison shows.} On the two camouflage prompted pools the candidate-derived DSS
adaptation and SPHERETRUST are within $0.1$ percentage points, $.7881$ against $.7873$ and $.8245$
against $.8241$, with the lower catastrophic-error rate on SPHERETRUST's side. They disagree on the
winner in $64\%$ of the merged camouflage pool's images, and on the salient, dichotomous and four
unprompted pools SPHERETRUST is ahead of the strongest evaluated external comparator by $1.7$ to
$9.3$ points.

\section{Candidate Set Training and the Second Round}
\label{app:cs}

\subsection{Ablation of the Training Method}
\label{app:csabl}

Tab.~\ref{tab:csabl} reports the training ablation with the seed standard deviations and the paired
difference of each variant against the full method per seed, three seeds being too few for an
interval. Every row trains the same architecture on the same images with the same schedule, only the
supervision changing. \emph{Selection only} trains on the mask of Eq.~(\ref{eq:R}) as a
fixed label, \emph{candidate set training} is round one of \S\ref{sec:m2}, and the rows below the
full method remove one part at a time. Without the completed mask, the round two set holds the five
leading candidates under $S$. Without cross fitting, $P_i$ comes from a single round one student
trained on all images. Without candidate set training, the final student trains on $U_i$ as a fixed
label. Without the sphere term in the round one order, round one orders its candidates by $\STscore$
alone, and without it in both rounds $S$ also drops $z(\Gamma)$. The two reference rows train on a
fixed label chosen by $\STscore+z(\Gamma)$ and on the oracle candidate. The full method exceeds
selection only and round one alone for every seed on every regime, the completed mask and cross
fitting provide the salient and dichotomous gains, candidate set training provides the camouflage
gain, and the sphere term contributes on camouflage and dichotomous data and is neutral on salient
data, although alone it reselects a better fixed label than $\STscore$ on every regime.

\begin{table}[!htb]
\caption{\textbf{Ablation of the training method}, $F^w_\beta$ on the MLLM anchor pools averaged over
every test set of the regime, mean $\pm$ population standard deviation over three seeds, the full
method in red, and the paired difference of each row against the full method in points for seeds
$0$ / $1$ / $2$. The seed varies the final student only. The fold split, the fold students, the
out-of-fold maps $P_i$, the prototype fit and the round two candidate sets were produced once with
seed $0$ and are shared, so these spreads describe the last stage conditional on that cache rather
than the end to end variance of the pipeline, which would be larger.}
\label{tab:csabl}
\apptab
\setlength{\tabcolsep}{1.6pt}
\begin{tabular}{@{}lcccccc@{}}
\toprule
& \multicolumn{2}{c}{COD} & \multicolumn{2}{c}{SOD} & \multicolumn{2}{c@{}}{DIS} \\
\cmidrule(lr){2-3}\cmidrule(lr){4-5}\cmidrule(l){6-7}
& $F^w_\beta$ & $\Delta$ per seed & $F^w_\beta$ & $\Delta$ per seed & $F^w_\beta$ & $\Delta$ per seed \\
\midrule
selection only (fixed rank one label) & .7513{\scriptsize$\pm$.0004} & $-4.3$/$-4.6$/$-4.6$ & .8148{\scriptsize$\pm$.0033} & $-2.7$/$-1.7$/$-2.5$ & .6048{\scriptsize$\pm$.0018} & $-5.7$/$-5.4$/$-5.3$ \\
$+$ candidate set training (round one) & .7930{\scriptsize$\pm$.0006} & $-0.3$/$-0.3$/$-0.3$ & .8232{\scriptsize$\pm$.0031} & $-1.3$/$-1.0$/$-2.0$ & .6384{\scriptsize$\pm$.0026} & $-1.9$/$-2.2$/$-2.3$ \\
$+$ second round (full method) & \best{.7961}{\scriptsize$\pm$.0007} &  & \best{.8376}{\scriptsize$\pm$.0013} &  & \best{.6594}{\scriptsize$\pm$.0014} &  \\
\multicolumn{7}{@{}l}{\emph{full method without}} \\
\quad completed mask & .7941{\scriptsize$\pm$.0002} & $-0.1$/$-0.2$/$-0.3$ & .8209{\scriptsize$\pm$.0010} & $-1.8$/$-1.4$/$-1.9$ & .6434{\scriptsize$\pm$.0010} & $-1.6$/$-1.6$/$-1.6$ \\
\quad cross fitting & .7953{\scriptsize$\pm$.0006} & $-0.1$/$-0.1$/$-0.1$ & .8280{\scriptsize$\pm$.0012} & $-0.9$/$-1.0$/$-1.0$ & .6475{\scriptsize$\pm$.0005} & $-1.3$/$-1.0$/$-1.3$ \\
\quad candidate set training & .7866{\scriptsize$\pm$.0007} & $-0.9$/$-1.0$/$-0.9$ & .8330{\scriptsize$\pm$.0030} & $-0.3$/$-0.7$/$-0.3$ & .6570{\scriptsize$\pm$.0002} & $-0.4$/$-0.1$/$-0.3$ \\
\quad sphere term, round one order & .7932{\scriptsize$\pm$.0015} & $-0.3$/$-0.1$/$-0.4$ & .8368{\scriptsize$\pm$.0030} & $-0.4$/$0.0$/$+0.2$ & .6579{\scriptsize$\pm$.0006} & $-0.3$/$0.0$/$-0.1$ \\
\quad sphere term, both rounds & .7896{\scriptsize$\pm$.0016} & $-0.8$/$-0.6$/$-0.6$ & .8377{\scriptsize$\pm$.0019} & $0.0$/$-0.1$/$+0.1$ & .6573{\scriptsize$\pm$.0008} & $-0.2$/$-0.1$/$-0.3$ \\
\midrule
sphere re-selected label, fixed & .7825{\scriptsize$\pm$.0010} &  & .8211{\scriptsize$\pm$.0015} &  & .6261{\scriptsize$\pm$.0008} &  \\
oracle candidate, fixed & .8048{\scriptsize$\pm$.0023} &  & .8311{\scriptsize$\pm$.0009} &  & .6745{\scriptsize$\pm$.0008} &  \\
\bottomrule
\end{tabular}
\end{table}

\PAR{What the second round costs and what it fixes.} Completion is a union, so it is
monotone in area, since it restores parts the candidates omit and leaves any background a candidate
leaked in place. What keeps the student off an over grown label is the candidate set, since a plain candidate
stays available at every step after the warm up. Subtracting a confident background region, which
would make the refinement bidirectional, remains unevaluated. Completion is worth $1.7$ and $1.6$ points on salient and dichotomous data, whose failures are
omitted parts, and $0.2$ on camouflage, whose failures are leaked background. The compute follows the
same split. The second round trains two extra students on half the data each, and on camouflage it
returns $0.3$ of the $4.5$ points, so round one alone buys $93\%$ of the gain for half the compute.
Ambiguous camouflaged targets rarely clear the fixed threshold $\tau=0.9$, so $U_i$ adds few pixels
over $M_i^S$ there. On salient and dichotomous data, where it returns $1.4$ and $2.1$ points, the
second round pays for itself.

\PAR{What the oracle row is.} The reference row \emph{oracle candidate, fixed} is a fixed label
student trained on the best single mask the generator's pool contains, so the full method exceeding
it on salient data exceeds a fixed label student rather than an annotation ceiling. The completed mask $U_i$ joins the
leading candidate with the confident region of a cross fitted prediction and can therefore cover parts
that no candidate in the pool covers, which moves the supervision outside the discrete candidate set
rather than above the ground truth.

\subsection{Choice of the Training Configuration}
\label{app:config}

Tab.~\ref{tab:config} lists every candidate configuration of the training stage with the validation
score that decided the choice (Appendix~\S\ref{app:impl}) and the test regime means of the same
runs. On the test sets the four completion thresholds lie within about one point of each other on
every regime, all four exceed round one alone on salient and dichotomous data and equal it on
camouflage, and the two orderings by $\STscore$ differ from the chosen one by at most $1.1$ points.

\begin{table}[!htb]
\caption{\textbf{Candidate configurations of the training stage.} Validation $F^w_\beta$ on the held
out split of each training pool with its mean over the three, the number the choice read, and the test
regime means of the same runs, single trainings with seed $0$ on the nine tenths of each pool outside
the validation split. The chosen configuration is in red.}
\label{tab:config}
\apptab
\setlength{\tabcolsep}{4pt}
\begin{tabular}{@{}lcccc ccc@{}}
\toprule
& \multicolumn{4}{c}{validation $F^w_\beta$} & \multicolumn{3}{c}{test regime mean $F^w_\beta$} \\
\cmidrule(lr){2-5}\cmidrule(lr){6-8}
configuration & COD & SOD & DIS & mean & COD & SOD & DIS \\
\midrule
\multicolumn{8}{@{}l}{\emph{round one}} \\
\quad order by $\STscore$, argmax rule & .7701 & .9377 & .7580 & .8219 & .7873 & .8210 & .6377 \\
\quad order by $\STscore$, softmax weights & .7675 & .9400 & .7549 & .8208 & .7841 & .8181 & .6371 \\
\quad order by $\STscore+z(\Gamma)$, argmax rule (chosen) & \best{.7759} & \best{.9392} & \best{.7544} & \best{.8232} & \best{.7955} & \best{.8185} & \best{.6372} \\
\multicolumn{8}{@{}l}{\emph{second round}} \\
\quad $U_i$ as a fixed label & .7713 & .9367 & .7775 & .8285 & .7878 & .8359 & .6505 \\
\quad candidate set without $U_i$ & .7751 & .9393 & .7566 & .8236 & .7922 & .8227 & .6388 \\
\quad candidate set with $U_i$, $\tau=0.7$ & .7753 & .9431 & .7844 & .8343 & .7937 & .8382 & .6558 \\
\quad candidate set with $U_i$, $\tau=0.8$ & .7766 & .9437 & .7865 & .8356 & .7953 & .8421 & .6566 \\
\quad candidate set with $U_i$, $\tau=0.9$ (chosen) & \best{.7776} & \best{.9433} & \best{.7836} & \best{.8348} & \best{.7946} & \best{.8324} & \best{.6551} \\
\quad candidate set with $U_i$, $\tau=0.95$ & .7756 & .9423 & .7793 & .8324 & .7945 & .8330 & .6557 \\
\bottomrule
\end{tabular}
\end{table}

\section{Downstream Results by Test Set}
\label{app:t2full}

Tabs.~\ref{tab:t2cod} to \ref{tab:t2dis} report the sets outside Tab.~\ref{tab:t2} with the metrics
customary for each regime. Rows marked $\dagger$ are transcribed from the source named in the caption,
and every other entry is evaluated over the complete set with our evaluator from released
predictions or weights (TSD and the full CSNet model released outputs for the salient sets only),
absent predictions count as empty maps (A2S-v3 on $3$ of $5{,}019$ DUTS-TE images, 3SD on $1$,
CutLER on $4$), and \best{red} and \second{blue} mark the best and second best per column, with our
rows shaded. Selfment
appears twice on camouflage, with the zero shot values its paper reports for a frozen DINOv3-7B
backbone and with our evaluation of its released weights, run as in its README in fp32 over the
complete set, on the five camouflage sets at the $1{,}536$ pixels that README uses by default and on
the salient and dichotomous sets at $768$, the resolution the released head was trained at, in an
earlier run whose maps we no longer hold. We could not reproduce the published COD10K result
under the stated released-checkpoint protocol: the published weighted $F$ is $.7540$ and ours at the
default resolution is $.5507$, a difference of $20.3$ percentage points, with $S_\alpha$ and
$E_\phi$ lower by $4.8$ and $5.6$ points and MAE higher by $2.5$ points. The published and reproduced results are labeled
separately. We scored the released maps with our evaluator and with Selfment's own \texttt{eval.py}
on the same predictions and ground truth, and the two agree to four decimals on all four metrics, so
the gap does not lie in evaluation. Checkpoint, inference entry point, resolution, probability
conversion, resampling, split and precision all match the release. Moving the five camouflage sets from $1{,}280$ to $1{,}536$ raises weighted $F$ by $0.1$
to $0.4$ points, so we report the released model as we measured it.

\PAR{Metric conventions.} Table~\ref{tab:t2sod} reports MAE, adaptive-threshold $F_\beta$, and
mean $E_\phi$. The transcribed CutLER result with $F_\beta$ at a fixed threshold of $0.5$ is
labeled accordingly and excluded from adaptive-threshold rankings. Table~\ref{tab:t2} reports
$S_\alpha$ and weighted $F$. Tab.~\ref{tab:t2sodsf} scores every evaluated row of Tab.~\ref{tab:t2sod} under
$S_\alpha$ and $F^w_\beta$ from the same scoring run, so the two conventions can be read side by
side, and their orders differ. Under MAE, $F_\beta$, and $E_\phi$, CSNet (full) leads our student on
DUTS-TE, HKU-IS, and DUT-OMRON. Under $S_\alpha$ and $F^w_\beta$ our student leads every evaluated
method on DUTS-TE, ECSSD, and PASCAL-S and on $S_\alpha$ of HKU-IS, CSNet (full) leads $F^w_\beta$
on HKU-IS, and A2S-v3 leads both on DUT-OMRON, where our student is fourth. Our students use the
MLLM anchor, and the camouflage table also lists the student of the pool SAM3 builds alone
(\S\ref{sec:e2}).

\begin{table}[!htb]
\caption{\textbf{Downstream, camouflage.} Rows marked $\dagger$ are transcribed, EASE (DINO ViT-S/8),
EReCu, FOUND, TokenCut, and UCOD-DPL (DINO ViT-S/8) from Table 1 of EReCu \citep{jiang2026erecu},
which reruns released code and copies EASE from its paper, the other EASE configurations from its
Tables 1 and 2, and RISE, SdalsNet, DualUCOD, DSS, UCOD-DPL (DINOv2), UCOD-MKD, and Selfment (DINOv3-7B)
from their papers, and \emph{Selfment, our evaluation} is its released checkpoint under our
evaluation. RISE uses SAM only to build its training masks, and SdalsNet reports max $F_\beta$ and
max $E_\phi$, so those entries stay empty. The rows below the last rule prompt SAM, HQ-SAM, or SAM2 at test
time, DSS also an MLLM, and are listed for reference outside the marks. Other rows use our evaluator on
released predictions.}
\label{tab:t2cod}
\apptab
\renewcommand{\arraystretch}{0.8}
\setlength{\tabcolsep}{1.9pt}
\begin{tabular}{@{}l rrrr rrrr rrrr@{}}
\toprule
& \multicolumn{4}{c}{CAMO, $n=250$} & \multicolumn{4}{c}{CHAMELEON, $n=76$} & \multicolumn{4}{c}{NC4K, $n=4121$} \\
\cmidrule(lr){2-5}\cmidrule(lr){6-9}\cmidrule(lr){10-13}
Method & $S_\alpha\uparrow$ & $F^w_\beta\uparrow$ & $E_\phi\uparrow$ & MAE$\downarrow$ & $S_\alpha\uparrow$ & $F^w_\beta\uparrow$ & $E_\phi\uparrow$ & MAE$\downarrow$ & $S_\alpha\uparrow$ & $F^w_\beta\uparrow$ & $E_\phi\uparrow$ & MAE$\downarrow$ \\
\midrule
EASE (DINO ViT-S/8)~\citep{du2025ease}$^\dagger$ & .6530 & .5630 & .7370 & .1660 & .6760 & .5500 & .7650 & .1050 & .7280 & .6330 & .7900 & .1080 \\
EASE (DINOv2 ViT-L/14)~\citep{du2025ease}$^\dagger$ & .7490 & .6840 & .8310 & .0980 & .8190 & .7410 & .8990 & .0440 & .8000 & .7350 & .8840 & .0560 \\
RISE (DINOv2 ViT-L/14)~\citep{du2025rise}$^\dagger$ & .7340 & .6100 & .7870 & .1090 & .8220 & .7200 & .8840 & .0500 & .8050 & .7050 & .8680 & .0610 \\
RISE (SAM ViT-H masks)~\citep{du2025rise}$^\dagger$ & .7600 & .6510 & .8070 & .1020 & .8230 & .7330 & .8820 & .0550 & .8250 & .7360 & .8740 & .0560 \\
DualUCOD (DINOv2)~\citep{liu2026dualucod}$^\dagger$ & .7710 & .7020 & .8520 & .0830 & .8270 & .7540 & .9180 & .0410 & .8080 & .7450 & .8960 & .0520 \\
SdalsNet (DINO ViT-S/8)~\citep{shou2025sdalsnet}$^\dagger$ & .6960 & -- & -- & .1170 & .7240 & -- & -- & .0800 & .7380 & -- & -- & .0840 \\
EReCu~\citep{jiang2026erecu}$^\dagger$ & .7027 & .6083 & .8003 & .1072 & .7321 & .6187 & .8523 & .0716 & .7583 & .6642 & .8498 & .0742 \\
FOUND~\citep{simeoni2023found}$^\dagger$ & .6913 & .6217 & .7465 & .1373 & .7161 & .6112 & .7704 & .0892 & .7459 & .6589 & .8073 & .0886 \\
TokenCut~\citep{wang2022tokencut}$^\dagger$ & .6450 & .5149 & .7237 & .1605 & .6573 & .5018 & .7351 & .1379 & .7338 & .6137 & .7919 & .1083 \\
UCOD-DPL (DINO ViT-S/8)~\citep{yan2025ucoddpl}$^\dagger$ & .7013 & .6109 & .7921 & .1082 & .7287 & .6154 & .8486 & .0725 & .7538 & .6674 & .8447 & .0745 \\
UCOD-DPL (DINOv2)~\citep{yan2025ucoddpl}$^\dagger$ & .7930 & .7470 & .8620 & .0770 & .8640 & \second{.8250} & \second{.9310} & .0310 & .8500 & .8180 & .9230 & .0430 \\
UCOD-MKD (PVTv2)~\citep{chen2026beyond}$^\dagger$ & .8090 & .7530 & .8750 & .0710 & -- & -- & -- & -- & .8570 & .8030 & .9180 & .0400 \\
UCOD-MKD (R50)~\citep{chen2026beyond}$^\dagger$ & .7640 & .7010 & .8330 & .0850 & -- & -- & -- & -- & .8210 & .7570 & .8840 & .0510 \\
Selfment, published~\citep{you2026selfment}$^\dagger$ & \best{.8690} & .7920 & .8940 & .0600 & \best{.9100} & \best{.8430} & \best{.9440} & \best{.0250} & \best{.9020} & \second{.8360} & \second{.9310} & \second{.0330} \\
\midrule
Selfment, our evaluation~\citep{you2026selfment} & .8633 & .6802 & .8406 & .0855 & \second{.8995} & .7234 & .8875 & .0469 & .8774 & .6789 & .8642 & .0634 \\
A2S-v3~\citep{yuan2024a2sv3} & .6889 & .5768 & .7543 & .1359 & .6710 & .5109 & .7328 & .1319 & .7412 & .6314 & .8109 & .0914 \\
CSNet-crf~\citep{guan2025csnet} & .6016 & .4300 & .6349 & .1479 & .5841 & .3354 & .5537 & .1010 & .6887 & .5480 & .7204 & .0895 \\
\midrule
\rowcolor{ourfill}
\textcolor{rkbest}{\textbf{SPHERETRUST}}, MLLM anchor & \second{.8640} & \best{.8121} & \best{.9089} & \second{.0531} & .8669 & .7937 & .9185 & .0344 & .8837 & .8340 & .9301 & .0350 \\
\rowcolor{ourfill}
\textcolor{rkbest}{\textbf{SPHERETRUST}}, SAM3 pool & .8558 & \second{.8034} & \second{.9034} & \best{.0527} & .8748 & .8102 & .9289 & \second{.0308} & \second{.8923} & \best{.8476} & \best{.9427} & \best{.0294} \\
\midrule
\multicolumn{13}{@{}l}{\emph{prompting SAM, HQ-SAM, SAM2, or an MLLM at test time}} \\
EASE $+$ SAM ViT-H~\citep{du2025ease}$^\dagger$ & .7770 & .7480 & .8410 & .0990 & .8530 & .8120 & .9120 & .0440 & .8490 & .8150 & .9020 & .0510 \\
EASE $+$ HQ-SAM ViT-H~\citep{du2025ease}$^\dagger$ & .8070 & .7710 & .8650 & .0780 & .8640 & .8270 & .9160 & .0370 & .8660 & .8330 & .9150 & .0390 \\
DSS (SAM2 ViT-L, MLLM)~\citep{yang2026dss}$^\dagger$ & .8080 & .7660 & .8700 & .0780 & .8830 & .8480 & .9260 & .0340 & .8910 & .8700 & .9400 & .0310 \\
\bottomrule
\end{tabular}
\end{table}

\begin{table}[!htb]
\caption{\textbf{Downstream, salient objects} (DUTS-TE $5{,}019$ images, ECSSD $1{,}000$, HKU-IS $4{,}447$,
DUT-OMRON $5{,}168$, PASCAL-S $850$). Rows marked $\dagger$ are transcribed, six methods from
the comparison table of A2S-v3 \citep{yuan2024a2sv3} and the DUTS-TE row of CutLER from Table I of
CSNet \citep{guan2025csnet}, each in its source's convention. The $F_\beta$ of that CutLER row,
marked $\ddagger$, is at a fixed threshold of $0.5$ and stays outside the adaptive-threshold
ranking. Evaluated rows use the adaptive threshold $F_\beta$ and the mean $E_\phi$, 3SD and UMNet
from their released maps (UMNet's release omits PASCAL-S), FOUND from its released weights (single
mask setting with the bilateral solver), and CutLER as the union of its released detector's instances at
confidence $0.35$, which differs from CSNet's CutLER row, so both are kept.}
\label{tab:t2sod}
\apptab
\renewcommand{\arraystretch}{0.84}
\setlength{\tabcolsep}{1.9pt}
\resizebox{\textwidth}{!}{\begin{tabular}{@{}l rrr rrr rrr rrr rrr@{}}
\toprule
& \multicolumn{3}{c}{DUTS-TE} & \multicolumn{3}{c}{ECSSD} & \multicolumn{3}{c}{HKU-IS} & \multicolumn{3}{c}{DUT-OMRON} & \multicolumn{3}{c}{PASCAL-S} \\
\cmidrule(lr){2-4}\cmidrule(lr){5-7}\cmidrule(lr){8-10}\cmidrule(lr){11-13}\cmidrule(lr){14-16}
Method & MAE$\downarrow$ & $F_\beta\uparrow$ & $E_\phi\uparrow$ & MAE$\downarrow$ & $F_\beta\uparrow$ & $E_\phi\uparrow$ & MAE$\downarrow$ & $F_\beta\uparrow$ & $E_\phi\uparrow$ & MAE$\downarrow$ & $F_\beta\uparrow$ & $E_\phi\uparrow$ & MAE$\downarrow$ & $F_\beta\uparrow$ & $E_\phi\uparrow$ \\
\midrule
A2S-v3~\citep{yuan2024a2sv3}$^\dagger$ & .0470 & .8160 & .9060 & .0380 & .9230 & \second{.9510} & .0330 & .9080 & \best{.9540} & .0620 & .7590 & \best{.8680} & .0690 & \second{.8440} & \second{.8990} \\
DCFD~\citep{lin2022dcfd}$^\dagger$ & .0640 & .7640 & .8550 & .0590 & .8880 & .9150 & .0420 & .8890 & .9350 & .0700 & .7100 & .8370 & .0900 & .7950 & .8600 \\
EDNS~\citep{zhang2020edns}$^\dagger$ & .0650 & .7350 & .8470 & .0680 & .8720 & .9060 & .0460 & .8740 & .9330 & .0760 & .6820 & .8210 & .0970 & .8010 & .8460 \\
STC~\citep{song2023stc}$^\dagger$ & .0520 & .8090 & .8910 & .0500 & .9030 & .9350 & .0410 & .8910 & .9420 & .0680 & .7530 & .8520 & .0760 & .8270 & .8810 \\
SelfMask~\citep{shin2022selfmask}$^\dagger$ & .0630 & .7140 & .8480 & .0580 & .8560 & .9200 & .0530 & .8190 & .9150 & .0780 & .6680 & .8150 & .0870 & .7740 & .8560 \\
TSD~\citep{zhou2023a2sv2}$^\dagger$ & .0470 & .8100 & .9010 & .0440 & .9160 & .9380 & .0370 & .9020 & .9470 & \second{.0610} & .7450 & .8630 & .0740 & .8300 & .8820 \\
CutLER~\citep{wang2023cutler}$^\dagger$ & .1310 & .6600$^{\ddagger}$ & .7670 & -- & -- & -- & -- & -- & -- & -- & -- & -- & -- & -- & -- \\
\midrule
3SD~\citep{yasarla20243sd} & .0869 & .6613 & .7859 & .0773 & .8370 & .8521 & .0676 & .8218 & .8533 & .0944 & .6475 & .7766 & .1188 & .7364 & .7875 \\
CutLER~\citep{wang2023cutler} & .2180 & .5442 & .6578 & .1263 & .7689 & .8175 & .1373 & .7253 & .7951 & .2592 & .4927 & .6053 & .1626 & .6899 & .7607 \\
FOUND~\citep{simeoni2023found} & .0579 & .7851 & .8709 & .0529 & .9088 & .9136 & .0630 & .8457 & .8651 & .0738 & .7106 & .8310 & .0790 & .8238 & .8740 \\
UMNet~\citep{wang2022umnet} & .0667 & .752 & .8448 & .0636 & .879 & .8989 & .0412 & .889 & .9267 & .0631 & .743 & .8392 & -- & -- & -- \\
Selfment, our evaluation~\citep{you2026selfment} & .1139 & .6936 & .7737 & .0994 & .8555 & .8251 & .0855 & .8400 & .8371 & .1730 & .5942 & .6695 & .1264 & .7599 & .7822 \\
A2S-v3~\citep{yuan2024a2sv3} & .0473 & .8154 & .9040 & \second{.0379} & \second{.9231} & .9478 & .0331 & .9080 & \second{.9498} & .0621 & \second{.7592} & \second{.8668} & \second{.0679} & .8380 & .8971 \\
CSNet-crf~\citep{guan2025csnet} & .0562 & .7784 & .8457 & .0540 & .8948 & .9086 & .0423 & .8866 & .9091 & .0808 & .6721 & .7673 & .0838 & .7998 & .8568 \\
CSNet (full)~\citep{guan2025csnet} & \best{.0431} & \best{.8515} & \best{.9110} & .0475 & .9121 & .9190 & \best{.0311} & \best{.9268} & .9447 & .0683 & \best{.7615} & .8409 & .0725 & .8380 & .8827 \\
TSD~\citep{zhou2023a2sv2} & .0468 & .8104 & .9019 & .0441 & .9167 & .9366 & .0365 & .9025 & .9424 & \best{.0609} & .7456 & .8635 & .0732 & .8260 & .8869 \\
\midrule
\rowcolor{ourfill}
\textcolor{rkbest}{\textbf{SPHERETRUST}}, MLLM anchor & \second{.0450} & \second{.8349} & \second{.9071} & \best{.0307} & \best{.9383} & \best{.9543} & \second{.0325} & \second{.9169} & .9435 & .0792 & .7271 & .8397 & \best{.0566} & \best{.8589} & \best{.9113} \\
\bottomrule
\end{tabular}}
\end{table}

\begin{table}[!htb]
\caption{\textbf{Downstream, salient objects, under $S_\alpha$ and $F^w_\beta$.} The evaluated rows of
Tab.~\ref{tab:t2sod} scored under the convention of Tab.~\ref{tab:t2} in the same scoring run, red
and blue marking the best and second best entry per column. Transcribed rows are left out since
their sources report neither.}
\label{tab:t2sodsf}
\apptab
\renewcommand{\arraystretch}{0.84}
\setlength{\tabcolsep}{3.4pt}
\begin{tabular}{@{}l rr rr rr rr rr@{}}
\toprule
& \multicolumn{2}{c}{DUTS-TE} & \multicolumn{2}{c}{ECSSD} & \multicolumn{2}{c}{HKU-IS} & \multicolumn{2}{c}{DUT-OMRON} & \multicolumn{2}{c@{}}{PASCAL-S} \\
\cmidrule(lr){2-3}\cmidrule(lr){4-5}\cmidrule(lr){6-7}\cmidrule(lr){8-9}\cmidrule(lr){10-11}
Method & $S_\alpha\uparrow$ & $F^w_\beta\uparrow$ & $S_\alpha\uparrow$ & $F^w_\beta\uparrow$ & $S_\alpha\uparrow$ & $F^w_\beta\uparrow$ & $S_\alpha\uparrow$ & $F^w_\beta\uparrow$ & $S_\alpha\uparrow$ & $F^w_\beta\uparrow$ \\
\midrule
3SD~\citep{yasarla20243sd} & .8121 & .6370 & .8848 & .7907 & .8774 & .7662 & .7979 & .6187 & .8060 & .6859 \\
CutLER~\citep{wang2023cutler} & .6503 & .5149 & .7897 & .7426 & .7686 & .7005 & .6092 & .4660 & .7275 & .6574 \\
FOUND~\citep{simeoni2023found} & .8029 & .7479 & .8691 & .8708 & .7959 & .7810 & .7709 & .6793 & .8075 & .7838 \\
UMNet~\citep{wang2022umnet} & .8027 & .7039 & .8677 & .8391 & .8865 & .8561 & .8047 & .6972 & -- & -- \\
Selfment~\citep{you2026selfment} & .7977 & .5659 & .8790 & .7194 & .8700 & .6966 & .7141 & .4284 & .8219 & .6500 \\
A2S-v3~\citep{yuan2024a2sv3} & .8477 & .7930 & \second{.9050} & \second{.9027} & .8977 & .8895 & \best{.8206} & \best{.7400} & \second{.8429} & \second{.8129} \\
CSNet-crf~\citep{guan2025csnet} & .7964 & .7275 & .8677 & .8582 & .8601 & .8470 & .7328 & .6151 & .8013 & .7589 \\
CSNet (full)~\citep{guan2025csnet} & \second{.8532} & \second{.8187} & .8862 & .8826 & \second{.9015} & \best{.9023} & .7944 & .7208 & .8305 & .8054 \\
TSD~\citep{zhou2023a2sv2} & .8424 & .7835 & .8935 & .8883 & .8899 & .8781 & \second{.8122} & \second{.7225} & .8301 & .7937 \\
\midrule
\rowcolor{ourfill}
\textcolor{rkbest}{\textbf{SPHERETRUST}}, MLLM anchor & \best{.8752} & \best{.8232} & \best{.9313} & \best{.9257} & \best{.9066} & \second{.8929} & .8018 & .7025 & \best{.8711} & \best{.8436} \\
\bottomrule
\end{tabular}
\end{table}

\begin{table}[!htb]
\caption{\textbf{Downstream, dichotomous segmentation}, marks per column within each set.}
\label{tab:t2dis}
\apptab
\renewcommand{\arraystretch}{0.84}
\setlength{\tabcolsep}{2.6pt}
\begin{minipage}[t]{0.49\textwidth}\centering
\resizebox{\linewidth}{!}{\begin{tabular}{@{}l rrrrr@{}}
\toprule
Method & $S_\alpha\uparrow$ & $F^w_\beta\uparrow$ & $E_\phi\uparrow$ & MAE$\downarrow$ & mDice$\uparrow$ \\
\midrule
\multicolumn{6}{@{}l@{}}{\textbf{DIS-TE1}, $n=500$} \\
Selfment~\citep{you2026selfment} & \second{.7054} & .3870 & .6628 & .1445 & \second{.5061} \\
A2S-v3~\citep{yuan2024a2sv3} & .6028 & \second{.4105} & \second{.6887} & .1379 & .4683 \\
CSNet-crf~\citep{guan2025csnet} & .5828 & .3625 & .6645 & \second{.1171} & .3996 \\
\rowcolor{ourfill}
\textcolor{rkbest}{\textbf{SPHERETRUST}}, MLLM anchor & \best{.7765} & \best{.6463} & \best{.8324} & \best{.0699} & \best{.6866} \\
\addlinespace[2pt]
\multicolumn{6}{@{}l@{}}{\textbf{DIS-TE2}, $n=500$} \\
Selfment~\citep{you2026selfment} & \second{.7573} & .4826 & .7091 & .1367 & \second{.5854} \\
A2S-v3~\citep{yuan2024a2sv3} & .6348 & \second{.4945} & \second{.7378} & .1424 & .5550 \\
CSNet-crf~\citep{guan2025csnet} & .6156 & .4629 & .7075 & \second{.1246} & .5015 \\
\rowcolor{ourfill}
\textcolor{rkbest}{\textbf{SPHERETRUST}}, MLLM anchor & \best{.7901} & \best{.6970} & \best{.8478} & \best{.0794} & \best{.7290} \\
\bottomrule
\end{tabular}}
\end{minipage}\hfill
\begin{minipage}[t]{0.49\textwidth}\centering
\resizebox{\linewidth}{!}{\begin{tabular}{@{}l rrrrr@{}}
\toprule
Method & $S_\alpha\uparrow$ & $F^w_\beta\uparrow$ & $E_\phi\uparrow$ & MAE$\downarrow$ & mDice$\uparrow$ \\
\midrule
\multicolumn{6}{@{}l@{}}{\textbf{DIS-TE3}, $n=500$} \\
Selfment~\citep{you2026selfment} & \second{.7640} & .4995 & .7110 & .1408 & .5850 \\
A2S-v3~\citep{yuan2024a2sv3} & .6586 & \second{.5403} & \second{.7736} & .1338 & \second{.5996} \\
CSNet-crf~\citep{guan2025csnet} & .6222 & .4831 & .7267 & \second{.1240} & .5176 \\
\rowcolor{ourfill}
\textcolor{rkbest}{\textbf{SPHERETRUST}}, MLLM anchor & \best{.7659} & \best{.6724} & \best{.8315} & \best{.0915} & \best{.7086} \\
\addlinespace[2pt]
\multicolumn{6}{@{}l@{}}{\textbf{DIS-TE4}, $n=500$} \\
Selfment~\citep{you2026selfment} & \best{.7372} & .4832 & .6864 & .1569 & .5495 \\
A2S-v3~\citep{yuan2024a2sv3} & .6247 & \second{.5160} & \second{.7420} & .1641 & \second{.5816} \\
CSNet-crf~\citep{guan2025csnet} & .5914 & .4534 & .6942 & \second{.1485} & .4909 \\
\rowcolor{ourfill}
\textcolor{rkbest}{\textbf{SPHERETRUST}}, MLLM anchor & \second{.7186} & \best{.6299} & \best{.7931} & \best{.1245} & \best{.6803} \\
\bottomrule
\end{tabular}}
\end{minipage}
\end{table}

\section{Fitted Rules and Other Generators}

\subsection{Fitted Scoring Rules and Evaluation Across Datasets}
\label{app:learned}

A linear predictor fitted to annotated mask quality is the natural alternative to equal weights,
with three features (the score terms) or eleven (every candidate statistic), scoring the same
candidates. Least squares on the P1 camouflage evaluation across four sets reaches selected Dice
$0.647$ with three features and $0.649$ with eleven, against $0.650$ for equal weights, and a separate
evaluation across four sets gives $0.457$ and $0.449$ against $0.478$. Weights fitted on one half of
the data and applied to the other give $0.617$, identical to equal weights, so we keep them.

\subsection{Other Generators and Anchors}
\label{app:anchor}
\label{app:nonsam}

\begin{table}[!htb]
\caption{\textbf{The rule across generators and anchors.} (a) The camouflage training pool from the
MLLM anchor with SAM and from SAM3 alone, pool oracle and selected Dice on the $2{,}770$ COD10K-train
images both pools contain, and the final student over four camouflage test sets and three seeds
(Appendix~\S\ref{app:t2full}). (b) Selection on a pool of $\nsNimg$ images from trained DPT-lite
decoders (Appendix~\S\ref{app:anchor}).}
\label{tab:pools}
\centering\footnotesize
\begin{minipage}[t]{0.62\textwidth}\centering
\textbf{(a) Camouflage training pool}\par\smallskip
\setlength{\tabcolsep}{3.2pt}
\resizebox{\linewidth}{!}{\begin{tabular}{@{}lcccccc@{}}
\toprule
& \multicolumn{2}{c}{pool} & \multicolumn{4}{c}{student, mean of four COD sets} \\
\cmidrule(lr){2-3}\cmidrule(lr){4-7}
pool & oracle & selected & $S_\alpha\uparrow$ & $F^w_\beta\uparrow$ & $E_\phi\uparrow$ & MAE$\downarrow$ \\
\midrule
MLLM anchor $+$ SAM & .858 & .790 & .8643 & .7961 & .9146 & .0395 \\
SAM3 alone & \textbf{.895} & \textbf{.829} & \textbf{.8710} & \textbf{.8086} & \textbf{.9257} & \textbf{.0342} \\
\bottomrule
\end{tabular}}
\end{minipage}\hfill
\begin{minipage}[t]{0.35\textwidth}\centering
\textbf{(b) Decoder pool}\par\smallskip
\setlength{\tabcolsep}{4pt}
\resizebox{\linewidth}{!}{\begin{tabular}{@{}lcc@{}}
\toprule
selector & sel.\ Dice & Top-1 \\
\midrule
generator confidence & \nsConf & .072 \\
$\Fcon+\Ebd$ pair $z(\Fcon)-z(\Ebd)$ & \nsHeu & .191 \\
SPHERETRUST & \textbf{\nsSt} & \textbf{.251} \\
oracle & \nsOrc & 1.000 \\
\bottomrule
\end{tabular}}
\end{minipage}
\end{table}

Here we build a pool from three trained DPT-lite decoders, sampled over checkpoints, four test time
views and three thresholds, deduplicated at IoU $>0.97$ and capped at twelve by
uniform sampling, a macro average over \nsNimg\ images from \nsNds\ sets outside the downstream
suite. The decoders were trained on masks selected by generator confidence, so only the candidate
generator differs from the SAM pools. Tab.~\ref{tab:pools}(b) reports the full score at \nsSt\
against \nsHeu\ for the $\Fcon+\Ebd$ pair and \nsConf\ for generator confidence (oracle \nsOrc,
random \nsRnd), a lead of \nsFinalDelta\ selected Dice over the pair (95\% CI
\nsFinalCi), with $\Fcon$ alone at $.775$, $\Csph$ alone at $.762$, and generator confidence keeping
the better AURC ($.126$ against $.218$).

\subsection{One Training Procedure, Three Selection Rules}
\label{app:armcmp}

Every downstream number elsewhere in the paper is supervised by SPHERETRUST's own ranking, so it
measures a pipeline rather than a selection rule. This experiment compares three
scoring rules under a common student architecture, training schedule, and candidate-set procedure.
The score transformations, candidate ordering, and training priors used for each rule are specified
below. On each of the three MLLM anchor training pools we run three rules,
SPHERETRUST, the size control $z(\Fcon)+z(\log a_{\mathrm{mask}})-z(\Ebd)$, and the candidate-derived
adaptation of the DSS scoring rule, each under two supervisions,
fixed label training on the mask that rule selects, and the complete two round procedure of
\S\ref{sec:m2} with the candidate sets ordered by that same rule, its own prototype fit, fold
students and completed masks. Images, student, schedule, hyperparameters and the three seeds are
identical across the six runs of a regime, and SPHERETRUST is retrained here rather than copied
from Tab.~\ref{tab:csabl}. The masks each rule selects reproduce its row of Tab.~\ref{tab:t1}, and
the rules disagree on $44\%$ to $66\%$ of the images of a pool, so the three carry different
supervision.

\PAR{Scores and priors of each rule.} Each alternative replaces $\STscore$ in Eq.~(\ref{eq:cs}) and
Eq.~(\ref{eq:S}) and nothing else changes. For SPHERETRUST, the size control, and adapted DSS, the
scores used for candidate ordering are $\rho+z_i(\Gamma)$ in round one and
$\rho+z_i(\Gamma)+z_i(\mathrm{IoU}(M,P_i))$ in round two, where $\rho$ is the rule itself,
$z(\Fcon)+z(\Csph)-z(\Ebd)$, $z(\Fcon)+z(\log a_{\mathrm{mask}})-z(\Ebd)$, or $z(s_{\mathrm{DSS}})$
with $s_{\mathrm{DSS}}$ the candidate-derived Eq.~(7) score, and every $z$ a standardization over
the candidates of one image. The priors supplied to Eq.~(\ref{eq:cs}) are these same ordering
scores, $\pi_{ik}$ equal to the round one score in round one and to the round two score in round
two, without further normalization or coefficients, with $T=0.1$, and the completed mask takes the
prior of the leading candidate. The rule term is standardized per term but not rescaled across
the three. Within an image $z(s_{\mathrm{DSS}})$ has unit variance, whereas the two sphere rules sum
three standardized terms, so the rule enters the ordering and the prior at a larger scale in those
two. The comparison uses a common student architecture and training schedule, with these stated
scoring and prior definitions.

\begin{table}[!htb]
\caption{\textbf{One training procedure, three scoring rules}, on the three MLLM anchor training
pools. The three rules are SPHERETRUST $z(\Fcon)+z(\Csph)-z(\Ebd)$, the size control
$z(\Fcon)+z(\log a_{\mathrm{mask}})-z(\Ebd)$, and the candidate-derived adaptation of the DSS
scoring rule. The selection columns report the mean Dice of each rule's selected masks. \emph{Fixed} and
\emph{full} report weighted $F$ for fixed label training and the complete procedure, as means $\pm$
population standard deviations across three final student seeds after averaging each seed's results
over the test sets of its regime. For the full procedure, these spreads describe final student
variation conditional on the shared upstream fold students, prototype fit, and round two candidate
cache of that rule. The student architecture, images, and schedule are matched, and the numerical
score definitions are specified above. SPHERETRUST was retrained for this comparison. Red
marks the highest observed mean, and paired per seed differences are reported for the full procedure
in Tab.~\ref{tab:armpair}.}
\label{tab:armcmp}
\apptab
\setlength{\tabcolsep}{3.4pt}
\begin{tabular}{@{}l rrr rrr rrr@{}}
\toprule
& \multicolumn{3}{c}{COD} & \multicolumn{3}{c}{SOD} & \multicolumn{3}{c@{}}{DIS} \\
\cmidrule(lr){2-4}\cmidrule(lr){5-7}\cmidrule(l){8-10}
ranking rule & sel & fixed & full & sel & fixed & full & sel & fixed & full \\
\midrule
SPHERETRUST & .787 & .7526{\scriptsize$\pm$.0004} & \best{.7971}{\scriptsize$\pm$.0009} & .929 & \best{.8144}{\scriptsize$\pm$.0006} & \best{.8403}{\scriptsize$\pm$.0009} & .771 & \best{.6047}{\scriptsize$\pm$.0013} & \best{.6597}{\scriptsize$\pm$.0010} \\
size control & .763 & .7287{\scriptsize$\pm$.0009} & .7958{\scriptsize$\pm$.0005} & .909 & .8009{\scriptsize$\pm$.0019} & .8367{\scriptsize$\pm$.0004} & .755 & .5842{\scriptsize$\pm$.0009} & .6571{\scriptsize$\pm$.0008} \\
adapted DSS & .788 & \best{.7624}{\scriptsize$\pm$.0011} & .7953{\scriptsize$\pm$.0016} & .912 & .7918{\scriptsize$\pm$.0012} & .8303{\scriptsize$\pm$.0018} & .720 & .5464{\scriptsize$\pm$.0032} & .6529{\scriptsize$\pm$.0011} \\
\bottomrule
\end{tabular}
\end{table}

\PAR{What it shows.} Under fixed label training the rules separate, by $3.4$ points of $F^w_\beta$
on camouflage, $2.3$ on salient objects and $5.8$ on dichotomous segmentation, and which rule comes
first depends on the pool. On camouflage the candidate-derived DSS adaptation trains the better student,
$.7624$ against $.7526$ for SPHERETRUST and $.7287$ for the size control, on every seed ($+0.9$ to
$+1.1$ points for DSS, $-2.3$ to $-2.5$ for the size control). The two leading rules select masks
of the same mean quality there, $.788$ against $.787$, so the difference lies in which masks they
select rather than in their average quality. On salient objects and on dichotomous segmentation
SPHERETRUST's ranking trains the better student, $.8144$ against $.7918$ for DSS and $.6047$
against $.5464$, by $2.1$ to $2.4$ and $5.6$ to $6.0$ points on every seed.
Under the complete procedure every rule gains, $4.5$, $2.6$ and $5.5$ points for SPHERETRUST,
$6.7$, $3.6$ and $7.3$ for the size control, and $3.3$, $3.9$ and $10.7$ for DSS, and the spread
among the three narrows to $0.2$, $1.0$ and $0.7$ points, with SPHERETRUST first on every regime
and every seed. The regime means lead the size control by $0.1$ to $0.4$ points and the adapted DSS
rule by $0.2$ to $1.0$, and the paired per seed differences of Tab.~\ref{tab:armpair} run from
$+0.1$ to $+0.5$ and from $+0.1$ to $+1.1$. In each regime the rule that starts furthest behind
gains the most. The candidate set, the prototypes and the
completion recover most of what the ranking rule decides, which is the sense in which the
downstream tables of this paper measure a pipeline rather than a selector.

The retrained SPHERETRUST runs reproduce the corresponding rows of Tab.~\ref{tab:csabl}, $.7513$,
$.8148$ and $.6048$ under fixed labels and $.7961$, $.8376$ and $.6594$ under the complete
procedure, to within $0.3$ points throughout.

\begin{table}[!htb]
\caption{\textbf{Paired per seed differences under the complete procedure}, SPHERETRUST minus the
named rule on the same seed, in points of $F^w_\beta$. Every difference is positive, so the ordering of
Tab.~\ref{tab:armcmp} holds seed by seed as well as in the mean.}
\label{tab:armpair}
\apptab
\setlength{\tabcolsep}{5pt}
\begin{tabular}{@{}l rrr rrr rrr@{}}
\toprule
& \multicolumn{3}{c}{COD} & \multicolumn{3}{c}{SOD} & \multicolumn{3}{c@{}}{DIS} \\
\cmidrule(lr){2-4}\cmidrule(lr){5-7}\cmidrule(l){8-10}
comparator & s0 & s1 & s2 & s0 & s1 & s2 & s0 & s1 & s2 \\
\midrule
size control & $+0.08$ & $+0.10$ & $+0.20$ & $+0.20$ & $+0.40$ & $+0.48$ & $+0.29$ & $+0.29$ & $+0.20$ \\
adapted DSS & $+0.26$ & $+0.19$ & $+0.09$ & $+1.13$ & $+1.02$ & $+0.88$ & $+0.94$ & $+0.66$ & $+0.45$ \\
\bottomrule
\end{tabular}
\end{table}

\PAR{Sensitivity to the score scale.} The three rule scores enter the candidate order and the prior
on different scales. Within an image the adapted DSS term is one standardized quantity of unit
variance, whereas each sphere rule sums three, and on $300$ images the variance of the SPHERETRUST
rule term is $3.56$. Standardizing each sphere rule once more, $z(\STscore)$ and
$z(z(\Fcon)+z(\log a_{\mathrm{mask}})-z(\Ebd))$, and substituting the standardized score for
$\rho$ in both rounds' ordering scores and priors puts the rule terms on the same within-image
scale as the adapted DSS term. This changes the rule term's weight relative to the prototype and
IoU terms, so the combined training orderings, retained candidate sets, and priors can change. Fixed label training is unaffected, since a positive affine map does not
move an argmax, and the complete procedure was retrained for both rules on all three pools.
Under the matched scale SPHERETRUST reaches $.7977$, $.8382$ and $.6593$ against $.7953$, $.8303$
and $.6529$ for the adapted DSS rule and $.7951$, $.8399$ and $.6602$ for the size control. The
margin over the adapted DSS rule survives on every regime and seed, from $+0.10$ to
$+1.02$ points, so it is not an artefact of the scale. The margin over the size control survives on
camouflage, from $+0.17$ to $+0.40$ points, and reverses on salient and dichotomous data by $0.17$
and $0.09$ points in the mean. Against the size control the comparison is therefore scale dependent
and the two rules are level, and the $0.1$ to $0.4$ points of Tab.~\ref{tab:armcmp} are not on
their own evidence that the spherical term beats the area surrogate.

\section{Four Further Benchmarks}
\label{app:fresh}

Tabs.~\ref{tab:freshsel} and \ref{tab:freshdown} score the rule, the generator's two scores,
SelfMask's consensus, SAQ, TokenCut, and the final students on four public test sets outside the $15$
of Appendix~\S\ref{app:regime2}, on prompted P2 pools built as in \S\ref{sec:exp}, with generator,
eligibility filter, selectors and students unchanged. The sets are the test split of PlantCamo \citep{yang2024plantcamo} ($250$
images) for camouflage, SOD300 \citep{movahedi2010sod} ($226$) and MSRA-B
\citep{liu2011msrab,jiang2013drfi} ($1{,}937$) for salient objects, and UHRSD \citep{xie2022uhrsd}
($988$) for dichotomous segmentation, after removal of the $74$ SOD300 and $63$ MSRA-B photographs
that ECSSD or the camouflage training split also contain.

\PAR{Selection.}
In Tab.~\ref{tab:freshsel}, the signed selected Dice margins over the strongest external comparator
are $-0.1$, $+1.8$, $+1.3$, and $+2.4$ percentage points on PlantCamo, SOD300, MSRA-B, and UHRSD,
respectively. The
candidate-derived DSS adaptation is that comparator on three of the four sets and is $0.1$ points
ahead on PlantCamo, where SPHERETRUST still leads it on Top-1. Coverage leads the size control by $.003$, $.014$, $.017$,
and $.017$, as on the prompted pools of Tab.~\ref{tab:t1}.

\PAR{Students.}
The final students exceed the selection only students on all four metrics on all four sets
($F^w_\beta$ $+1.6$, $+5.5$, $+1.4$, $+4.0$), and on
PlantCamo the SAM3 pool student trails the MLLM anchor student by $9.3$ points of $F^w_\beta$ and
falls $7.7$ below even the selection only student. This difference may partly reflect the animal-focused phrase ladder used to build
the SAM3 training pool (\emph{animal, camouflaged animal, insect, object}), whereas PlantCamo's
targets are plants and the MLLM anchor prompt names no taxon. The comparison does not isolate the
effect of the phrase ladder from other differences between the two training pools.

\begin{table}[!htb]
\caption{\textbf{Unsupervised selection on the four further benchmarks}, prompted pools, $n$
eligible images, $k$ candidates per image, selected Dice. Red and blue mark the highest and
second highest point estimates among the evaluated external selectors and SPHERETRUST, below them
the size control and the oracle.}
\label{tab:freshsel}
\apptab
\renewcommand{\arraystretch}{0.84}
\setlength{\tabcolsep}{9pt}
\begin{tabular}{@{}lcccc@{}}
\toprule
Selector & \textbf{camouflage} & \multicolumn{2}{c}{\textbf{salient}} & \multicolumn{1}{c@{}}{\textbf{dichotomous}} \\
\multicolumn{1}{@{}l}{} & \itshape PlantCamo & \itshape SOD300 & \itshape MSRA-B & \itshape UHRSD \\
\multicolumn{1}{@{}l}{} & \itshape $n{=}233$, $k{=}34.7$ & \itshape $219$, $29.7$ & \itshape $1932$, $24.6$ & \itshape $969$, $26.8$ \\
\midrule
Random & .434 & .555 & .747 & .676 \\
SelfMask vote & .505 & .611 & .873 & .789 \\
SAQ & .433 & .519 & .769 & .683 \\
SAM predicted IoU & .474 & .625 & .856 & .770 \\
SAM stability & .459 & .612 & .849 & .758 \\
TokenCut & .560 & \second{.681} & .868 & .814 \\
Generator confidence with UCOD-MKD-inspired filtering & .485 & .626 & .875 & .791 \\
Adapted DSS scoring rule (candidate-derived map) & \best{.569} & .668 & \second{.889} & \second{.821} \\
\rowcolor{ourfill}
\textcolor{rkbest}{\textbf{SPHERETRUST}} & \second{.568} & \best{.699} & \best{.902} & \best{.845} \\
\midrule
\emph{size control} & .565 & .685 & .885 & .828 \\
oracle & .661 & .760 & .932 & .891 \\
\bottomrule
\end{tabular}
\end{table}

\begin{table}[!htb]
\caption{\textbf{Students on the four further benchmarks}, mean of three seeds (seed standard
deviations at most $.016$), the full method against selection only.}
\label{tab:freshdown}
\apptab
\renewcommand{\arraystretch}{0.9}
\setlength{\tabcolsep}{5pt}
\begin{tabular}{@{}l rrrr@{}}
\toprule
student & $S_\alpha\uparrow$ & $F^w_\beta\uparrow$ & $E_\phi\uparrow$ & MAE$\downarrow$ \\
\midrule
\multicolumn{5}{@{}l}{\emph{camouflage students}, PlantCamo-TE} \\
full method, MLLM anchor & .686 & .515 & .763 & .097 \\
full method, SAM3 pool & .621 & .422 & .682 & .110 \\
selection only, MLLM anchor & .681 & .499 & .755 & .101 \\
\multicolumn{5}{@{}l}{\emph{salient student}, SOD300} \\
full method & .765 & .736 & .797 & .116 \\
selection only & .730 & .681 & .752 & .131 \\
\bottomrule
\end{tabular}\hspace{14pt}
\begin{tabular}{@{}l rrrr@{}}
\toprule
student & $S_\alpha\uparrow$ & $F^w_\beta\uparrow$ & $E_\phi\uparrow$ & MAE$\downarrow$ \\
\midrule
\multicolumn{5}{@{}l}{\emph{salient student}, MSRA-B-TE} \\
full method & .906 & .883 & .930 & .045 \\
selection only & .898 & .869 & .921 & .048 \\
\multicolumn{5}{@{}l}{\emph{dichotomous student}, UHRSD-TE} \\
full method & .885 & .850 & .908 & .052 \\
selection only & .859 & .810 & .873 & .071 \\
& & & & \\
\bottomrule
\end{tabular}
\end{table}

}

\end{document}